%% file: main.tex
\documentclass{article}
\usepackage[margin=1.2in]{geometry}

\usepackage{lineno}

\usepackage{setspace}

\usepackage{enumitem}
\usepackage{graphicx}
\usepackage[english]{babel}
\usepackage{amsmath,amsfonts, amsthm}
\usepackage{bm}
\usepackage{bbm}

\usepackage{subfig,balance}

\usepackage{cases}
\usepackage[overload]{empheq}

\usepackage{wrapfig}

\usepackage{caption}

\usepackage{color}

\usepackage[dvipsnames,svgnames,x11names,hyperref]{xcolor}

\usepackage{algorithm}
\usepackage{algorithmic}

\usepackage{amssymb}

\usepackage{mathtools}

\usepackage{hyperref}
\hypersetup{colorlinks,breaklinks,      citecolor=ForestGreen,urlcolor=NavyBlue, linkcolor=NavyBlue}

\usepackage[nameinlink]{cleveref}

\allowdisplaybreaks

\input{AMS_commands.tex}

\input{tex_macros}

\title{\vspace{-2em}
\textbf{High Dimensional Robust $M$-Estimation:  Arbitrary Corruption and Heavy Tails}}

\author{
 Liu Liu \\
   \texttt{liuliu@utexas.edu}
  \and
Tianyang Li\\
   \texttt{lty@cs.utexas.edu}
   \and
Constantine Caramanis \\
\texttt{constantine@utexas.edu}
\and
   The University of Texas at Austin
}

\date{}

\begin{document}

\maketitle

\begin{abstract}
We consider the problem of sparsity-constrained $M$-estimation when both {\em explanatory and response} variables have heavy tails (bounded 4-th moments), or a fraction of arbitrary corruptions. We focus on the $k$-sparse, high-dimensional regime where the number of variables $d$ and the sample size $n$ are related through $n \sim k \log d$. We define a natural condition we call the Robust Descent Condition (RDC), and show that if a gradient estimator satisfies the RDC, then Robust Hard Thresholding (IHT using this gradient estimator), is guaranteed to obtain good statistical rates. The contribution of this paper is in showing that this RDC is a flexible enough concept to recover known results, and obtain new robustness results. Specifically, new results include: (a) For $k$-sparse high-dimensional linear- and logistic-regression with heavy tail (bounded 4-th moment) explanatory and response variables, a linear-time-computable median-of-means gradient estimator satisfies the RDC, and hence Robust Hard Thresholding is minimax optimal; (b) When instead of heavy tails we have $O(1/\sqrt{k}\log(nd))$-fraction of arbitrary corruptions in explanatory and response variables, a near linear-time computable trimmed gradient estimator satisfies the RDC, and hence Robust Hard Thresholding is minimax optimal.
We demonstrate the effectiveness of our approach in sparse linear, logistic regression, and sparse precision matrix estimation on synthetic and real-world US equities data.
\end{abstract}


\input{Introduction}

\input{RobustGradient}

\input{Theory}

\input{Graphical.tex}

\input{Experiments.tex}

\clearpage

{
\small
\bibliography{Notes_Robust}
\bibliographystyle{alpha}
}

\clearpage

\appendix

\input{Proofs.tex}

\input{Additional.tex}

\end{document}

%% file: AMS_commands.tex
\theoremstyle{plain}

\newtheorem{theo}{Theorem}[section]

\newtheorem{lem}{Lemma}[section]
\newtheorem{prop}{Proposition}[section]
\newtheorem{cor}{Corollary}[section]

\newtheorem{nota}{Notation}[section]
\newtheorem{de}{Definition}[section]
\newtheorem{model}{Model}[section]

\newtheorem{exa}{Example}[section]
\newtheorem{as}{Assumption}[section]
\newtheorem{alg}{Algorithm}[section]

\newcommand{\btheo}{\begin{theo}}
\newcommand{\bde}{\begin{de}}
\newcommand{\ble}{\begin{lem}}
\newcommand{\bpr}{\begin{prop}}
\newcommand{\bno}{\begin{nota}}
\newcommand{\bex}{\begin{exa}}
\newcommand{\bcor}{\begin{cor}}
\newcommand{\spro}{\begin{proof}}
\newcommand{\bas}{\begin{as}}
\newcommand{\balg}{\begin{alg}}

\newcommand{\etheo}{\end{theo}}
\newcommand{\ede}{\end{de}}
\newcommand{\ele}{\end{lem}}
\newcommand{\epr}{\end{prop}}
\newcommand{\eno}{\end{nota}}
\newcommand{\eex}{\end{exa}}
\newcommand{\ecor}{\end{cor}}
\newcommand{\fpro}{\end{proof}}
\newcommand{\eas}{\end{as}}
\newcommand{\ealg}{\end{alg}}

\theoremstyle{plain}

\newtheorem{theos}{Theorem}
\newtheorem{props}{Proposition}
\newtheorem{lems}{Lemma}
\newtheorem{cors}{Corollary}

\newtheorem{exas}{Example}
\newtheorem{algs}{Algorithm}
\newtheorem{asss}{Assumption}
\newtheorem{defns}{Definition}

\newcommand{\btheos}{\begin{theos}}
\newcommand{\etheos}{\end{theos}}
\newcommand{\bprops}{\begin{props}}
\newcommand{\eprops}{\end{props}}
\newcommand{\bdes}{\begin{defns}}
\newcommand{\edes}{\end{defns}}
\newcommand{\blems}{\begin{lems}}
\newcommand{\elems}{\end{lems}}
\newcommand{\bcors}{\begin{cors}}
\newcommand{\ecors}{\end{cors}}
\newcommand{\bexs}{\begin{exas}}
\newcommand{\eexs}{\end{exas}}
\newcommand{\balgs}{\begin{algs}}
\newcommand{\ealgs}{\end{algs}}
\newcommand{\bass}{\begin{asss}}
\newcommand{\eass}{\end{asss}}

%% file: tex_macros.tex
\DeclareMathOperator{\Expe}{\mathbb{E}}

\DeclareMathOperator{\Real}{\mathbb{R}}
\DeclareMathOperator{\Var}{\mathrm{Var}}
\DeclareMathOperator{\Cov}{\mathrm{Cov}}
\DeclareMathOperator*{\argmin}{arg\,min}

\makeatletter
\newcommand*\rel@kern[1]{\kern#1\dimexpr\macc@kerna}
\newcommand*\widebar[1]{%
  \begingroup
  \def\mathaccent##1##2{%
    \rel@kern{0.8}%
    \overline{\rel@kern{-0.8}\macc@nucleus\rel@kern{0.2}}%
    \rel@kern{-0.2}%
  }%
  \macc@depth\@ne
  \let\math@bgroup\@empty \let\math@egroup\macc@set@skewchar
  \mathsurround\z@ \frozen@everymath{\mathgroup\macc@group\relax}%
  \macc@set@skewchar\relax
  \let\mathaccentV\macc@nested@a
  \macc@nested@a\relax111{#1}%
  \endgroup
}
\makeatother

\newcommand{\eps}{\epsilon}

\DeclarePairedDelimiter\abs{\lvert}{\rvert}%
\DeclarePairedDelimiter\norm{\lVert}{\rVert}%

\makeatletter
\let\oldabs\abs
\def\abs{\@ifstar{\oldabs}{\oldabs*}}
\let\oldnorm\norm
\def\norm{\@ifstar{\oldnorm}{\oldnorm*}}
\makeatother

\newcommand{\parambar}{\widebar{\param}}
\newcommand{\Deltabar}{\widebar{\Delta}}

\newcommand{\parambestk}{\param^*}

\newcommand{\z}{\bm{z}}

\newcommand{\x}{\bm{x}}

\newcommand{\constraint}{B}

\newcommand{\supp}{\mathrm{supp}}

\newcommand{\sconvexity}{\mu_\alpha}
\newcommand{\ssmoothness}{\mu_L}
\newcommand{\slack}{c_\rho}
\newcommand{\condnum}{\rho}

\newcommand{\loss}{\ell}

\newcommand{\inner}[2]{\left\langle {#1} , {#2}  \right\rangle}
\newcommand{\smallinner}[2]{\langle {#1} , {#2} \rangle}

\newcommand{\UMean}{{\bm{G}}}

\newcommand{\OGood}{{\mathcal{G}}}
\newcommand{\OGrad}{\UMean}
\newcommand{\obj}{f}

\newcommand{\RobustG}{\widehat{\bm{G}}}

\newcommand{\Good}{\widetilde{\mathcal{G}}}
\newcommand{\Bad}{\mathcal{B}}

\newcommand{\Remain}{\mathcal{R}}
\newcommand{\Trim}{\mathcal{T}}

\newcommand{\Input}{\mathcal{S}}

\newcommand{\trim}{{\sf trmean}}
\newcommand{\Hard}[1]{P_{#1}}

\newcommand{\thr}{\Hard{k'}}

\newcommand{\gradsample}{{\bm {g}}}
\newcommand{\vect}{{\bm {v}}}

\newcommand{\param}{{\bm{\beta}}}

\newcommand{\NeiRV}{\bm{X}_{(j)}}

\newcommand{\NeiCov}{\bm{\Sigma}_{(j)}}

\newcommand{\NeiVec}{\bm{\sigma}_{(j)}}

\newcommand{\precision}{\bm{\Theta}}
\newcommand{\Sig}{\bm{\Sigma}}

\newcommand{\opnorm}[1]{\left\lVert#1\right\rVert_{\rm op}}



%% file: Introduction.tex
\section{Introduction}
\label{sec:intro}


$M$-estimation is a standard technique for statistical estimation \cite{van2000asymptotic}. The past decade has seen successful extensions of $M$-estimation to the high dimensional setting with sparsity (or other low-dimensional structure), e.g., using Lasso \cite{tibshirani1996regression,buhlmann2011statistics,hastie2015statistical,WainwrightBook}.
Yet sparse modeling in high dimensions is NP-hard in the worst case \cite{Bandeira2013Certifying,zhang2014lower}. Thus theoretical sparse recovery guarantees for most computationally tractable approaches (e.g., $\ell_1$ minimization \cite{donoho2006compressed,Cands2004RobustUP,wainwright2009sharp}, Iterative Hard Thresholding \cite{IHT2009}) rely on 
strong assumptions on the probabilistic models of the data, such as sub-Gaussianity. Under such assumptions, these approaches achieve the minimax rate for sparse regression \cite{raskutti2011minimax}.

Meanwhile, statistical estimation with heavy tailed outliers or even arbitrary corruptions has long been a focus in robust statistics \cite{box1953non,tukey1975mathematics,huber2011robust,hampel2011robust}.\footnote{Following \cite{minsker2018AOS,fan2016shrinkage}, by {\em heavy-tail} we mean satisfying only weak moment bounds, specifically, bounded 4-th order moments (compared to sub-exponential or sub-Gaussian).} But heavy-tails and arbitrary corruptions in the data violate the assumptions required for convergence of the usual algorithms. A central question then, is what assumptions are sufficient to enable efficient and robust algorithms for high dimensional $M$-estimation under heavy tails or arbitrary corruption.

Huber's seminal work \cite{huber1964robust} and more modern followup work \cite{loh2017statistical} has considered replacing the classical least squared risk minimization objective with a robust counterpart (e.g., Huber loss). Other approaches (e.g., \cite{li2013compressed}) considered regularization-based robustness approaches.
However, when there are outliers in the explanatory variables (covariates), these approaches do not seem to succeed \cite{chen2013robust}. Meanwhile, approaches combining recent advances in robust mean estimation and gradient descent have proved remarkably powerful in the low-dimensional setting \cite{ravikumar2018robust,klivans2018efficient,sever2018}, but for high dimensions, have so far only managed to address the setting where the covariance of the explanatory variables is the identity, or sparse \cite{du2017computationally,liu2018high}. Meanwhile, flexible and statistically optimal approaches (\cite{gao2017robust}) have relied on intractable estimators such as Tukey-depth.

For the heavy-tail setting, another line of research considers estimators such as Median-of-Means (MOM) \cite{MOM_nemirovsky1983problem,MOM_jerrum1986random,MOM_alon1999space,minsker2015geometric} and Catoni's mean estimator \cite{catoni2012poincare,minsker2018AOS} only use weak moment assumptions. \cite{minsker2015geometric, brownlees2015empirical, hsu2016loss}
generalized these ideas to $M$-estimation, yet it is not clear if these approaches apply to the high-dimensional setting with heavy tailed covariates. 

{\bf Main Contributions.} In this paper, we develop a sufficient condition that when satisfied, guarantees that an efficient algorithm (a variant of IHT) achieves the minimax optimal statistical rate. We show that our condition is flexible enough to apply to a number of important high-dimensional estimation problems under either heavy tails, or arbitrary corruption of the data. Specifically:


\begin{enumerate}[leftmargin=*]
\item We consider two models. For our arbitrary corruption model, we assume that an adversary {\em replaces} an arbitrary $\epsilon$-fraction of the authentic samples with arbitrary values (\Cref{def:contamination_model}). For the heavy-tailed model, we assume our data (response and covariates) satisfy only weak moment assumptions (\Cref{def:heavy_model}) without sub-Gaussian or sub-exponential concentration bounds.

\item  We propose a notion that we call the Robust Descent Condition (RDC). 
Given any gradient estimator that satisfies the RDC, we define RHT -- Robust Hard Thresholding (\Cref{alg:THT}) for sparsity constrained $M$-estimation, and prove that \Cref{alg:THT} converges linearly to a minimax statistically optimal solution. Thus the RDC and Robust Hard Thresholding form the basis for a Deterministic Meta-Theorem (\Cref{thm:meta}) that guarantees estimation error rates as soon as the RDC property of any gradient estimator can be certified.


\item We then obtain non-asymptotic bounds via certifying the RDC for different robust gradient estimators under various statistical models. (A) For corruptions in both response and explanatory variables, we show the trimmed gradient estimator satisfies the RDC. Thus our algorithm RHT has minimax-optimal statistical error, and tolerates   
$O({1}/{(\sqrt{k} \log (nd))})$-fraction of outliers. This fraction is nearly independent of the $d$, which is important in the high dimension regime.     
(B) In the heavy tailed regime, we use the Median-of-Means (MOM) gradient estimator.
Our RHT algorithm obtains the sharpest available error bound, in fact nearly matching the results in the sub-Gaussian case. With either of these gradient estimators, our algorithm is computationally efficient, nearly matching vanilla  gradient descent. This is in particular much faster than algorithms relying on sparse PCA relaxations as subroutines (\cite{du2017computationally,liu2018high}).

\item  We use Robust Hard Thresholding for neighborhood selection \cite{meinshausen2006high} for estimating Gaussian graphical models, and provide model selection guarantees under adversarial corruption of the data; our results share similar robustness guarantees with sparse regression.

\item We demonstrate the effectiveness of Robust Hard Thresholding on both arbitrarily corrupted/heavy tailed synthetic data and (unmodified) real data.
\end{enumerate}
A concrete illustration of 3(B) above: Consider a sparse linear regression problem without noise (sparse linear equations), with scaling $n = {O}(k \log d)$. When the covariates are sub-Gaussian, Lasso succeeds in exact recovery with high probability (as expected). When the covariates have only 4-th moments, we do not expect Lasso to succeed, and indeed experiments indicate this. Moreover, to the best of our knowledge, no previous efficient algorithm with ${O}(k\log(d))$ samples can guarantee exact recovery in this observation model (\cite{fan2016shrinkage} has a statistical rate depending on the norm of the parameter $\bm{\beta}^{\ast}$, and thus exact recovery for $\sigma = 0$ is not guaranteed). Our contributions show that Robust Hard Thresholding using MOM achieves this (see also simulations in  \Cref{fig:Linear}(b)).

\subsection*{Related work}


{\bf Sparse regression with  arbitrary corruptions or heavy tails.} Several works in robustness of high dimensional problems consider heavy tailed distributions or arbitrary corruptions only in the response variables \cite{li2013compressed,bhatia2015robust,bhatia2017consistent,loh2017statistical,price2018compressed, hsu2016loss,minsker2015geometric,chimedian}.
Yet these algorithms cannot be trivially extended to the setting with heavy tails or corruptions in explanatory variables. 
Another line  \cite{alfons2013sparse,vainsencher2017ignoring,Trim2018general,shen2018iteratively} focuses on alternating minimization approaches which extend Least Trimmed Squares \cite{rousseeuw1984least}. However, these methods only have local convergence guarantees, and cannot handle arbitrary corruptions.

\cite{chen2013robust} was one of the first papers to provide guarantees for sparse regression with arbitrary outliers in both response and explanatory variables by trimming the design matrix.
Similar trimming techniques are also used in \cite{fan2016shrinkage} for heavy tails in response and explanatory variables.
Those results are specific to sparse regression, however, and cannot be easily extended to general $M$-estimation problems. Moreover, even for linear regression, the statistical rates are not minimax optimal.
\cite{lugosi2016tourament}
uses Median-of-Means tournaments to deal with heavy tails in the explanatory variables and obtains near optimal rates. However, 
Median-of-Means tournaments is not known to be computationally tractable. \cite{lecue2017robust}
deals with heavy tails and outliers in the explanatory variables, but they require higher moment bound (whose order is $O(\log(d))$) in the isotropic design case.
\cite{gao2017robust} optimizes Tukey depth \cite{tukey1975mathematics,gao2018aos} for
robust sparse regression under the Huber $\eps$-contamination model, and their algorithm is minimax optimal and can handle a constant fraction of outliers. However, computing Tukey depth is intractable \cite{johnson1978densest}. Recent results \cite{du2017computationally, liu2018high} leverage robust sparse mean estimation in robust sparse regression.
Their algorithms are computationally tractable, and can tolerate $\eps = \text{const.}$, but they require very restrictive assumptions on the covariance matrix ($\bm{\Sigma} = \bm{I}_d$ or sparse), which precludes their use in applications such as graphical model estimation.

{\bf Robust $M$-estimation via robust gradient descent.}
Works in \cite{chen2017distributed,holland2017efficient} and later \cite{Yin_median} first leveraged the idea of using robust mean estimation in each step of gradient descent, using a subroutine such as geometric median.
A similar approach using more sophisticated robust mean estimation methods was later proposed in \cite{ravikumar2018robust,sever2018,Yin_nonconvex, su2018securing,holland2018AiStats} for robust gradient descent. These methods all focused on low dimensional robust $M$-estimation.  Work in \cite{liu2018high} extended the approach to the high-dimensional setting (though is limited to $\bm{\Sigma} = \bm{I}_d$ or sparse covariances).
Even though the corrupted fraction $\eps$ can be independent of the ambient dimension $d$ by using sophisticated robust mean estimation algorithms
\cite{Moitra2016FOCS,Lai2016FOCS,Steinhardt2017Arxiv}, or the sum-of-squares framework \cite{klivans2018efficient}, these algorithms (except \cite{liu2018high}) are not applicable to the high dimensional setting ($n \ll d$), as they require
at least $\Omega(d)$ samples.

{\bf Robust estimation of graphical models.}
A line of research using a robustified covariance matrix
in Gaussian graphical models \cite{liu2012Skeptic, gu2017robust, loh2018Spearman}
leverages GLasso \cite{friedman2008sparse} or CLIME \cite{cai2011constrained} to estimate the sparse precision matrix. These robust methods are
restricted to Gaussian graphical model estimation, and their techniques cannot be generalized to
other $M$-estimation problems.

{\bf Notation.}
We denote the Hard Thresholding operator of sparsity $k'$ by $\Hard{k'}$,
and denote the Euclidean projection onto the $\ell_2$ ball $\constraint$ by $\Pi_\constraint$.
We use $\Expe_{i\in_u \Input}$ to denote the expectation operator obtained by the uniform
distribution over all samples $\{i\in \Input\}$.


\section{Problem formulation}
\label{sec:setup}
We now define the corruption and heavy tails model and sparsity constrained $M$-estimation.
\bde[$\eps$-corrupted samples]
\label{def:contamination_model}
Let $\{\bm{z}_i, i\in \OGood\}$ be i.i.d. observations with     
distribution $P$. We say that a collection of samples $\{\bm{z}_i, i\in\Input\}$ is $\eps$-corrupted if an adversary chooses an  arbitrary $\eps$-fraction of the samples in $\OGood$ and modifies them with arbitrary values.
\ede
This corruption model allows
corruptions in both {\em explanatory and response} variables in regression problems where we observe $\bm{z}_i = (y_i, \bm{x}_i)$.
\Cref{def:contamination_model} also allows the adversary to select an
$\eps$-fraction of samples to \emph{delete and corrupt}.
\bde[heavy-tailed samples] 
\label{def:heavy_model}
For a distribution $P$ of $\x \in \Real^d$ 
with mean $\Expe(\x)$ and covariance $\Sig$,
we say that $P$ has bounded $2k$-th moment,
if there is a universal constant $C_{2k}$ such that, for a unit vector $\vect \in \Real^d$, we have
$\Expe_P \abs{\inner{\vect}{\x-\Expe(\x)}}^{2k} \leq C_{2k}
\Expe_P(\abs{\inner{\vect}{\x-\Expe(\x)}}^{2})^k$.
\ede
\Cref{def:heavy_model} allows heavy tails in both {\em explanatory and response} variables for $\bm{z}_i = (y_i, \bm{x}_i)$.
For example, in \Cref{thm:main_linear_heavy}, we study linear regression with bounded 4-th moments for $\x$ and bounded variance for $y$ and noise.

Let $\ell : \Real^d \times \mathcal{Z} \rightarrow \Real$ be a convex and differentiable loss function.
Our target is the unknown sparse
population minimizer $\param^* = \arg\min_{\param \in \Real^d, \norm{\param}_0 \leq k} \Expe_{\bm{z}_i \sim P} \loss_i(\param; \bm{z}_i)$, and
we write $\obj$ as the population risk,
$\obj(\param) = \Expe_{\bm{z}_i \sim P} \loss_i(\param; \bm{z}_i)$.
Note that $\param^*$'s definition allows model misspecification.
The following \Cref{def:population risk} provides general assumptions for the population risk.


\bde[Strong convexity/smoothness]\label{def:population risk}
For the population risk $\obj$, we assume 
$\sconvexity \norm*{\param_1 - \param_2}_2^2/2 \leq \obj(\param_1) - \obj(\param_2) - \inner{\nabla\obj(\param_2)}{\param_1 - \param_2} \leq \ssmoothness\norm*{\param_1 - \param_2}_2^2/2$,
where $\sconvexity$ is the strong-convexity parameter  and
$\ssmoothness$ is the smoothness parameter.
The condition number is $\condnum = {\ssmoothness}/{\sconvexity} \geq 1$.
\ede

A well known result \cite{negahban2012unified}
considers ERM with convex relaxation from $\norm{\param}_0$ to $\norm{\param}_1$, by certifying the RSC condition for sub-Gaussian ensembles 
--
this obtains uniform convergence of the empirical risk.
From an optimization viewpoint,
existing results reveal that gradient descent algorithms equipped with soft-thresholding \cite{agarwal2012}
or hard-thresholding
\cite{IHT2009,jain2014iterative,Ping2017tight, GradientHT2018, barber2018between} have
linear convergence rate, and achieve
known
minimax lower bounds
in statistical estimation \cite{raskutti2011minimax,zhang2014lower}.


Given samples $\Input$,
running ERM on the entire input dataset:
$\min_{\param \in \constraint, \norm*{\param}_0 \leq k } \Expe_{i \in_u \Input} \loss_i(\param; \bm{z}_i)$,
cannot guarantee uniform convergence of the empirical risk, and can be arbitrarily bad
for $\eps$-corrupted samples. The next two sections outline the main results of this paper, addressing this problem.


%% file: RobustGradient.tex
\section{Robust sparse estimation
via Robust Hard Thresholding}
\label{sec:THT}

We introduce our meta-algorithm, Robust Hard Thresholding, that essentially uses a robust gradient estimator to run IHT. We require several definitions to specify the algorithm, and describe its results. 
We use $\RobustG(\param)$ as a placeholder for the estimate at $\param$, obtained from whichever robust gradient estimator we are using. Let $\UMean(\param) = \Expe_{\bm{z}_i \sim P} \nabla\loss_i(\param; \bm{z}_i)$ denote the population gradient. We use $\RobustG$ and $\UMean$ when the context is clear. 

Many previous works (\cite{chen2017distributed,holland2017efficient, ravikumar2018robust, sever2018, Yin_median, Yin_nonconvex, su2018securing}) have provided algorithms for obtaining robust gradient estimators, then used as subroutines in robust gradient algorithms. However, those results require controlling $\| \RobustG - \UMean\|_2$, and do not readily extend to high dimensions, as sufficiently controlling $\| \RobustG - \UMean\|_2$ seems to require $n = \Omega(d)$. 
A recent work \cite{liu2018high} on robust sparse linear regression uses a robust sparse mean estimator \cite{du2017computationally} to guarantee  $\norm*{\RobustG - \UMean}_2 = O(\delta_1 \norm{{\param} - \param^*}_2 +  \delta_2)$ with sample complexity $\Omega(k^2 \log(d))$. However, their algorithm requires the restrictive assumption $\bm{\Sigma} = I_d$ or sparse, and thus cannot be extended to more general $M$-estimation problems.

To address this issue, we
propose Robust Hard Thresholding
(\Cref{alg:THT}),
which uses hard thresholding after each robust gradient update\footnote{Our theory requires splitting samples across different iterations to maintain independence between iterations. We believe this is an artifact of the analysis, and do not use this  in our experiments. \cite{balakrishnan2017statistical, ravikumar2018robust} use a similar approach for theoretical analysis. }. In line 7, we use a gradient estimator to obtain the robust gradient estimate $\RobustG^t$. In line 8, we update the parameter by hard thresholding $\parambar^{t+1} = \Hard{k'}(\param^{t} - \eta \RobustG^t)$, where the hyper-parameter $k'$  proportional to $k$ is specified in \Cref{def:population risk}. A key observation in line 8 is that, in each step of IHT, the iterate $\param^t$ is sparse, and thus the \emph{perturbation from outliers or heavy tails} only depends on IHT's sparsity $k'$ instead of the ambient dimension $d$. Based on a careful analysis of the hard thresholding operator in each iteration, we show that rather than controlling $\norm*{\RobustG - \UMean}_2$, it is enough to control a weaker quantity: this is what we call the Robust Descent Condition \Cref{def:RDC} and we define it next; it  plays a key role in obtaining \emph{sharp} rates of convergence for various types of statistical models.

\begin{algorithm}[t]
\begin{algorithmic}[1]
\STATE \textbf{Input:} Data samples $\{y_i, \bm{x}_i\}_{i=1}^N$, gradient estimator $\RobustG$.
\STATE \textbf{Output:} The estimation $\widehat{\param}$.
\STATE \textbf{Parameters:} Hard thresholding parameter $k' = 4\condnum^2 k$.\\
{\kern2pt \hrule \kern2pt}
\STATE Split samples into $T$ subsets each of size $n$.  Initialize with $\param^0 = \bm{0}_d$.
\FOR {$t=0$ to $T-1$,}
\STATE At current $\param^t$, calculate all gradients for current $n$ samples:
$\gradsample_i^t = \nabla \loss_i(\param^t)$, $i\in [n]$.
\STATE For $\{\gradsample_i^t\}_{i=1}^n$, we obtain $\RobustG^t$ 

 

\STATE Update the parameter:
\label{line:update}
$\parambar^{t+1} = \Hard{k'}\Big(\param^{t} - \eta \RobustG^t\Big).$
Then project:
$\param^{t+1} = \Pi_\constraint (\parambar^{t+1})$.
\ENDFOR
\STATE Output the estimation $\widehat{\param} = \param^{T}$.
\end{algorithmic}
\caption{Robust Hard Thresholding}
\label{alg:THT}
\end{algorithm}

\subsection*{Robust Descent Condition}

The Robust Descent Condition \cref{equ:RDC} provides an upper bound on the inner product of the robust gradient estimate and the distance to the population optimum. This is a natural notion to control the  potential progress obtained by using a robust gradient update instead of the population gradient.

\bde[$(\alpha, \psi)$-Robust Descent Condition (RDC)]
\label{def:RDC}
For the population gradient ${\UMean}$ at $\param$,
a robust gradient estimator
$\RobustG(\param)$ satisfies
the robust descent condition
if for any sparse $\param, \widetilde{\param} \in \Real^d$,
\begin{align}
\label{equ:RDC}
    \abs{ \inner{\RobustG(\param) - {\UMean}(\param)}{\widetilde{\param} - \param^*}} \leq
   \Big ( \alpha \norm{\param - \param^*}_2 + \psi  \Big) \norm{\widetilde{\param} - \param^*}_2.
\end{align}
\ede

We begin with a Meta-Theorem for \Cref{alg:THT} that holds 
under the Robust Descent Condition \Cref{def:RDC} and assumptions on population risk \Cref{def:population risk}.
In \Cref{thm:meta}, we prove \Cref{alg:THT}'s global convergence and its statistical guarantees.
The proofs are collected in \Cref{sec:proof_stat}.

\btheo[Meta-Theorem]
\label{thm:meta}
Suppose we observe samples from a statistical model with population risk $f$ satisfying \Cref{def:population risk}.
If a robust gradient estimator satisfies $(\alpha, \psi)$-Robust Descent Condition (\Cref{def:RDC})
where $\alpha \leq \frac{1}{32}\sconvexity$, then
\Cref{alg:THT} with $\eta = 1/\ssmoothness$ outputs $\widehat{{\param}}$ such that
$\norm*{\widehat{{\param}} - \param^*}_2
= O(\psi/\sconvexity)$,
by setting
$ T = {O}\left(
\condnum \log\left( \sconvexity{\norm{\param^*}_2}/
\psi \right)\right)$.
\etheo

We note that \Cref{thm:meta} is deterministic in nature. 
In the sequel, we omit the log term in the sample complexity due to sample splitting.
We obtain  high probability results via certifying that the RDC holds for certain robust gradient estimators under various statistical models. To obtain the minimax estimation error rate in \Cref{thm:meta}, the key step is providing a robust gradient estimator with sufficiently small $\psi$, in the definition of RDC. 

\Cref{sec:new_section} uses the RDC and \Cref{thm:meta} to obtain new results for sparse regression under heavy tails or arbitrary corruption. Before we move to this, we observe that we can use the RDC and \Cref{thm:meta} to recover existing results in the literature. Some immediate examples are as follows:

\textbf{Uncorrupted gradient satisfies the RDC. }
Suppose the samples follow from sparse linear regression with sub-Gaussian covariates and noise $\mathcal{N}(0, \sigma^2)$. The empirical average of gradient samples satisfies \cref{equ:RDC} with $\psi = O(\sigma\sqrt{k\log(d)/n})$, by assuming $\ell_1$ constraint on ${\param}$ and $\widetilde{\param}$ \cite{loh2011high}. Plugging in this $\psi$ to \Cref{thm:meta} recovers the well-known minimax rates for sparse linear regression \cite{raskutti2011minimax}.

\textbf{RSGE implies RDC. }
When $\bm{\Sigma} = \bm{I}_d$ or is sparse,
\cite{du2017computationally} and \cite{liu2018high}, respectively, provide robust sparse gradient estimators (RSGE) which upper bound $\norm*{\RobustG(\param) - {\UMean}(\param)}_2 \leq \alpha \norm{\param - \param^*}_2 + \psi$, for  a constant fraction $\epsilon$ of corrupted samples. Noting that $\abs*{\langle{\RobustG(\param) - {\UMean}(\param)}, {\widetilde{\param} - \param^*}\rangle} \leq \norm*{\RobustG(\param) - {\UMean}(\param)}_2 \norm*{\widetilde{\param} - \param^*}_2$, we observe that RSGE implies RDC. Hence any RSGE can be used in \Cref{alg:THT}. The RSGE for $\Sig = I$ in \cite{du2017computationally} guarantees an RDC with $\psi = O(\sigma \epsilon)$ when $n = \Omega(k^2 \log d /\eps^2)$, and the RSGE for unknown sparse $\Sig$ from \cite{liu2018high} guarantees $\psi = O(\sigma \sqrt{\epsilon})$ when $n = \Omega(k^2 \log d /\eps)$. Again plugging these values for $\psi$ into our theorem, recovers the results in those papers. \footnote{It remains an open question to obtain a RSGE for a constant fraction of outliers for robust sparse regression with arbitrary covariance $\Sig$.}

\section{Main Results: Using the RDC and \Cref{alg:THT}}

\label{sec:new_section}
In the remainder of our paper, we use \Cref{thm:meta} and the RDC to analyze two well-known and computationally efficient robust mean estimation subroutines that have been used in the low-dimensional setting: the trimmed mean estimator and the MOM estimator. 
We show that these two can  obtain a sufficiently small $\psi$ in the definition of the  RDC. This leads to the minimax estimation error in the case of arbitrary corruptions or heavy tails.

\subsection{  
Gradient estimation}
\label{sec:grad_est}

The trimmed mean and MOM estimators have been successfully applied to robustify gradient descent 
\cite{Yin_median,ravikumar2018robust} in the low dimensional setting. They have not been used in the high dimensional regime, however, because until now we have not had the machinery to analyze their algorithmic convergence, statistical rates and minimax optimality in the high dimensional setting. 

To show they satisfy  the RDC with a sufficiently small $\psi$, we observe that by using H\"older's inequality on the LHS of \cref{equ:RDC}, we have $\abs*{ \smallinner{\RobustG(\param) - {\UMean}(\param)}{\widetilde{\param} - \param^*}} \leq \norm*{{\RobustG(\param) - {\UMean}(\param)}}_\infty \norm*{\widetilde{\param} - \param^*}_1.$
Using \Cref{alg:THT}, the Hard Thresholding step enforces sparsity of $\widetilde{\param} - \param^*$. Therefore, controlling $\psi$ amounts to bounding the infinity norm of the robust gradient estimate. 

In \Cref{sec:guarantees}, we show that by using coordinate-wise robust mean estimation, we can certify the RDC with sufficiently small $\psi$ to guarantee minimax rates. Specifically, we show this for the trimmed gradient estimator for arbitrary corruption, and and the MOM gradient estimator for heavy tailed distributions.

\bde\label{def:Robust_Gradient}
Given  gradients samples $\{\nabla\loss_i(\param; \bm{z}_i)\in \Real^d, i\in\Input\}$,
for each dimension $j \in [d]$,

$(\spadesuit)$: \textbf{Trimmed gradient estimator} removes the
largest and smallest $\alpha$ fraction of elements in
$\{ [\nabla\loss_i(\param; \bm{z}_i)]_j \in \Real, i\in\Input\}$, and calculates the mean of the remaining terms.
We choose $\alpha = c_0\eps$
for constant $c_0\geq 1$,
and require
$\alpha \leq 1/2 - c_1$ for a small constant $c_1 > 0$.

$(\clubsuit)$: \textbf{MOM gradient estimator} partitions $\Input$ into 
$ 4.5 \lceil\log(d)\rceil$ blocks and
    computes the sample mean of $\{ [\nabla\loss_i(\param; \bm{z}_i)]_j \in \Real\}$ within each block,  and then take the median of these means.\footnote{Without loss of generality, we assume the number of blocks divides $n$, and $ 4.5 \lceil\log(d)\rceil$ is chosen in  \cite{hsu2016loss}.}
\ede

%% file: Theory.tex
\subsection{Statistical guarantees}
\label{sec:guarantees}

In this section,
we consider some typical models for general $M$-estimation.

\begin{model}[Sparse linear regression]
\label{model:linear}
Samples $\bm{z}_i = (y_i, \bm{x}_i)$ are drawn from a linear model $P$:
$y_i = \bm{x}_i^{\top}\param^* + \xi_i$,
with $\param^*\in \Real^d$ being  $k$-sparse.
We assume that $\bm{x}$'s are  i.i.d. with normalized covariance matrix $\bm{\Sigma}$, with $\bm{\Sigma}_{jj}\leq 1$ $\forall j$, and the stochastic noise
$\xi$ has mean $0$ and variance $\sigma^2$.
\end{model}

\begin{model}[Sparse logistic regression]
\label{model:logistic}
Samples
$\bm{z}_i = (y_i, \bm{x}_i)$ are drawn from a binary classification model $P$, where
the binary label
$y_i\in \{-1, +1\}$ follows the conditional probability distribution
$\Pr(y_i|\bm{x}_i) = {1}/({1 + \exp(-y_i \bm{x}_i^\top \param^*)})$,
with $\param^* \in \constraint \subset \Real^d$ being $k$-sparse.
We assume that $\bm{x}$'s are i.i.d. with normalized covariance matrix $\bm{\Sigma}$, where $\bm{\Sigma}_{jj}\leq 1$ for all $j$.
\end{model}

To obtain the following corollaries, we first certify the RDC for a certain robust gradient estimator
over random ensembles with corruption or heavy tails, and then use them in \Cref{thm:meta}.
We collect the results for gradient estimation in \Cref{sec:proof_grad}, and the proofs for corollaries  in \Cref{sec:proof_stat}.

\paragraph{Arbitrary corruption case.}


Based on \Cref{thm:meta}, we first provide concrete results for arbitrary corruption case \Cref{def:contamination_model},
where the covariates and response variables in the authentic distribution $P$ are assumed to be sub-Gaussian.
%

\bcor\label{thm:main_linear}
Suppose we observe $n$ $\eps$-corrupted (\Cref{def:contamination_model}) sub-Gaussian samples from sparse linear regression model (\Cref{model:linear}).
Under the condition  $n = \Omega\left({ \condnum^4 k \log d} \right)$, and $\eps =  O\Bigl(\frac{1}{ \condnum^2 \sqrt{k} \log (nd)}\Bigr)$,
with probability at least $1-d^{-2}$, \Cref{alg:THT} with trimmed gradient estimator satisfies the RDC with $\psi = O(\condnum \sigma \sqrt{k} ({ \eps \log (nd) } +\sqrt{{\log d}/{n}}))$,
and thus \Cref{thm:meta} provides
$
\norm*{\widehat{{\param}} - \param^*}_2 = O ( { \condnum^2 \sigma } ( {\eps \sqrt{k} \log (nd) } + {\sqrt{{k \log d}/{n}}} )).
$
\ecor

\textit{Time complexity.}
\Cref{thm:main_linear} has a global linear convergence rate. In each iteration, we only use $O(n d \log n)$ operations complexity to calculate trimmed mean.
We incur logarithmic overhead       
compared to normal gradient descent \cite{bubeck2015convex}.

\textit{Statistical accuracy and robustness.}
Compared with \cite{chen2013robust, du2017computationally}, our statistical error rate is minimax optimal \cite{raskutti2011minimax,zhang2014lower}, and has no dependencies on $\norm{\param^*}_2$.
Furthermore, the upper bound on $\eps$ is nearly independent of  $d$, which guarantees \Cref{alg:THT}'s robustness in high dimensions.

\bcor\label{thm:main_logistic}
Suppose we observe $n$  $\eps$-corrupted (\Cref{def:contamination_model}) sub-Gaussian samples from sparse logistic regression model (\Cref{model:logistic}).
With probability at least $1-d^{-2}$,
\Cref{alg:THT} with trimmed gradient estimator satisfies the RDC with
$\psi = O(\condnum \sqrt{k} ({ \eps \log (nd) } +\sqrt{{\log d}/{n}}))$,
and thus
\Cref{thm:meta} provides
$\norm*{\widehat{{\param}} - \param^*}_2
= O( { \condnum }^2
(
{\eps \sqrt{k} \log (nd) } +
{\sqrt{{k \log d}/{n}}}
))$.
\ecor
\textit{Statistical accuracy and robustness.}
Under the sparse Gaussian linear discriminant analysis model (a typical example of \Cref{model:logistic}),
\Cref{alg:THT} achieves the statistical minimax rate
\cite{li2015fast, li2017minimax}.

\paragraph{Heavy-tailed distribution case.}

%
We next turn to the heavy tailed distribution case \Cref{def:heavy_model}. 

\bcor\label{thm:main_linear_heavy}
Suppose we observe $n$  samples from sparse linear regression model (\Cref{model:logistic}) with bounded 4-th moment covariates.
Under the condition  $n = \Omega\left({ \condnum^6 k \log d} \right)$,
with probability at least $1-d^{-2}$, \Cref{alg:THT} with MOM gradient estimator satisfies the RDC with $ \psi = O
( { \condnum^{3/2} \sigma  }
{\sqrt{{k \log d}/{n}}}
)$,
and thus \Cref{thm:meta} provides
$
\norm*{\widehat{{\param}} - \param^*}_2
= O
( { \condnum^{5/2} \sigma  }
{\sqrt{{k \log d}/{n}}}
)$.
\ecor

\textit{Time complexity.}
Similar to \Cref{thm:main_linear}, \Cref{thm:main_linear_heavy} has a global linear convergence. In each iteration, we only use $O(n d)$ operations complexity --  the same as normal gradient descent \cite{bubeck2015convex}.

\textit{Statistical accuracy.}
\cite{lugosi2016tourament} uses Median-of-Means tournaments to deal with sparse linear regression with bounded moment assumptions for the covariates, and they obtain near optimal rates. We obtain similar rates, however our algorithm is efficient, where as Median-of-Means tournaments is not known to be computationally tractable.  
\cite{fan2016shrinkage,zhu2017taming} deal with the same problem by truncating and shrinking the data to certify the RSC condition. Their results require boundedness of higher moments of the noise $\xi$, and the final error depends on $\norm{\param^*}_2$.
Our estimation error bounds exactly recover optimal sub-Gaussian bounds for sparse regression \cite{negahban2012unified,WainwrightBook}, and moreover, we obtain exact recovery when $\xi$'s variance $\sigma^2 \rightarrow 0$. 


\bcor\label{thm:main_logistic_heavy}
Suppose we observe $n$  samples from sparse logistic regression model (\Cref{model:logistic}).
With probability at least $1-d^{-2}$, \Cref{alg:THT} with MOM gradient estimator satisfies the RDC with $ \psi = O
( { \condnum^{3/2} }
{\sqrt{{k \log d}/{n}}}
)$,
and thus \Cref{thm:meta} provides
$
\norm*{\widehat{{\param}} - \param^*}_2
= O
( { \condnum^{5/2} }
{\sqrt{{k \log d}/{n}}}
)$.
\ecor

%% file: Graphical.tex
\subsection{Sparsity recovery and Gaussian graphical model estimation}
\label{sec:sparsity}
We next demonstrate the sparsity
recovery performance of \Cref{alg:THT} for  graphical model
learning \cite{meinshausen2006high, wainwright2009sharp,ravikumar2010ising, ravikumar2011logdet,buhlmann2011statistics,hastie2015statistical}. Our sparsity recovery guarantees hold for both heavy tails and arbitrary corruption, though we only present  results in the case of arbitrary corruption in this section.


We use $\supp(\vect, k)$ to denote top $k$ indexes of $\vect$ with the largest magnitude.    
Let $\vect_\mathrm{min}$ denote the smallest absolute value of nonzero element of $\vect$.
To control the false negative rate,
\Cref{thm:sparsity} shows that    
under the $\param_\mathrm{min}$-condition,    
$\supp(\widehat{\param}, k)$  is exactly $\supp({\param^*})$.    
The proofs are given
in \Cref{sec:appendix_sparsity}.
Sparsity recovery guarantee for sparse logistic regression is similar, and is omitted due to space constraints.
Existing results
on sparsity recovery for $\ell_1$ regularized estimators
\cite{wainwright2009sharp, ravikumar2015sparsistency}
do not require the RSC condition,
but instead require an irrepresentability condition,
which is stronger.
If $\eps\rightarrow 0$,
\Cref{thm:sparsity} has the same $\param_\mathrm{min}$-condition as IHT for sparsity recovery   
\cite{GradientHT2018}.

\bcor
\label{thm:sparsity}
Under the same condition as in \Cref{thm:main_linear}, and
a $\param_\mathrm{min}$-condition    on $\param^*$, 
$\param_\mathrm{min}^* =
    \Omega( { \condnum^2 \sigma }
(
{ \eps \sqrt{k} \log (nd) } +
\sqrt{{k \log d}/{n}} ))$,
 \Cref{alg:THT} with trimmed gradient estimator 
guarantees that
$\supp(\widehat{\param}, k) = \supp({\param^*})$, with  probability   at   least $1 - d^{-2}$.
\ecor


We consider sparse precision matrix estimation for Gaussian graphical models. The sparsity pattern of its precision matrix  $\precision = \bm{\Sigma}^{-1}$ matches the conditional independence relationships \cite{koller2009probabilistic, wainwright2008graphical}.
\begin{model}[Sparse precision matrix estimation]
\label{model:precision}
Under the contamination model \Cref{def:contamination_model},
authentic samples
$\{\bm{x}_i\}_{i=1}^m$ are drawn
from a multivariate Gaussian distribution $\mathcal{N}(0, \bm{\Sigma})$.
We assume that   
each row of the precision matrix $\precision = \bm{\Sigma}^{-1}$ is $(k+1)$-sparse -- each node has at most $k$ edges.
\end{model}

For the uncorrupted samples drawn from the  Gaussian graphical model,
the neighborhood
selection (NS) algorithm  \cite{meinshausen2006high}  solves a   
convex relaxation of the following sparsity constrained optimization to
regress each variable
against its neighbors
\begin{align}
     \widehat{\param}_j = \argmin_{\param \in\Real^{d-1}} \frac{1}{m} \sum_{i=1}^m (x_{ij} - \bm{x}_{i(j)}^\top\param)^2, \quad
      \text{ s.t. }
   \norm{\param}_0 \leq k, \quad\quad \text{for each } j\in[d], \label{equ:NS}
\end{align}
where $x_{ij}$ denotes the $j$-th coordinate of $x_i\in \Real^d$, and $(j)$ denotes the
index set $\{1, \cdots , j- 1, j + 1, \cdots , d\}$. 
Let $\bm{\theta}_{(j)} \in \Real^{d-1}$  denote $\precision$'s $j$-th column with the diagonal entry removed.
and $\bm{\Theta}_{j,j} \in \Real$  denote the $j$-th diagonal element of $\bm{\Theta}$.
Then, the sparsity pattern of $\bm{\theta}_{(j)}$
can be estimated through $\widehat{\param}_j$.
Details on the connection between $\bm{\theta}_{(j)}$  and $\widehat{\param}_j$ are given in \Cref{sec:appendix_sparsity}.

However, given $\eps$-corrupted samples from the Gaussian graphical model,
this procedure will fail \cite{liu2012Skeptic,gu2017robust}.   
To address this issue, we propose Robust NS
(\Cref{alg:graphical_IHT} in \Cref{sec:appendix_sparsity}), which
robustifies Neighborhood Selection \cite{meinshausen2006high} by
using Robust Hard Thresholding (with least square loss) to robustify \cref{equ:NS}.
Similar to  \Cref{thm:sparsity},
a $\bm{\theta}_\mathrm{min}$-condition guarantees consistent edge selection.
\bcor\label{cor:sparsity}
Under the same condition as in \Cref{thm:main_linear}, and
a $\bm{\theta}_\mathrm{min}$-condition for $\bm{\theta}_{(j)}$,
$\bm{\theta}_{(j),\mathrm{min}} = \Omega ( { {\precision_{j,j}^{1/2}} \condnum^2 }({ \eps \sqrt{k} \log (nd) } +
\sqrt{{k \log d}/{n}} ))$,
Robust NS (\Cref{alg:graphical_IHT})  achieves consistent
edge selection, with probability at least $1 - d^{-1}$.
\ecor

Similar to \Cref{thm:main_linear}, the fraction $\eps$ is nearly independent of dimension $d$, which provides guarantees of Robust NS in high dimensions.
Other Gaussian graphical model selection algorithms include GLasso \cite{friedman2008sparse}, CLIME\cite{cai2011constrained}.
The experimental details comparing robustified versions of these algorithms are presented in \Cref{sec:Huge}.


%% file: Experiments.tex
\section{Experiments}
\label{sec:experiments}

We provide the complete details for our experiment setup in \Cref{sec:additional}. 

\begin{figure}
\centering
\begin{minipage}{.48\textwidth}
\centering
\subfloat[][$\log(\norm{{\param^t} -\param^*}_2)$  vs. iterates.]{
\includegraphics[width=.47\columnwidth]{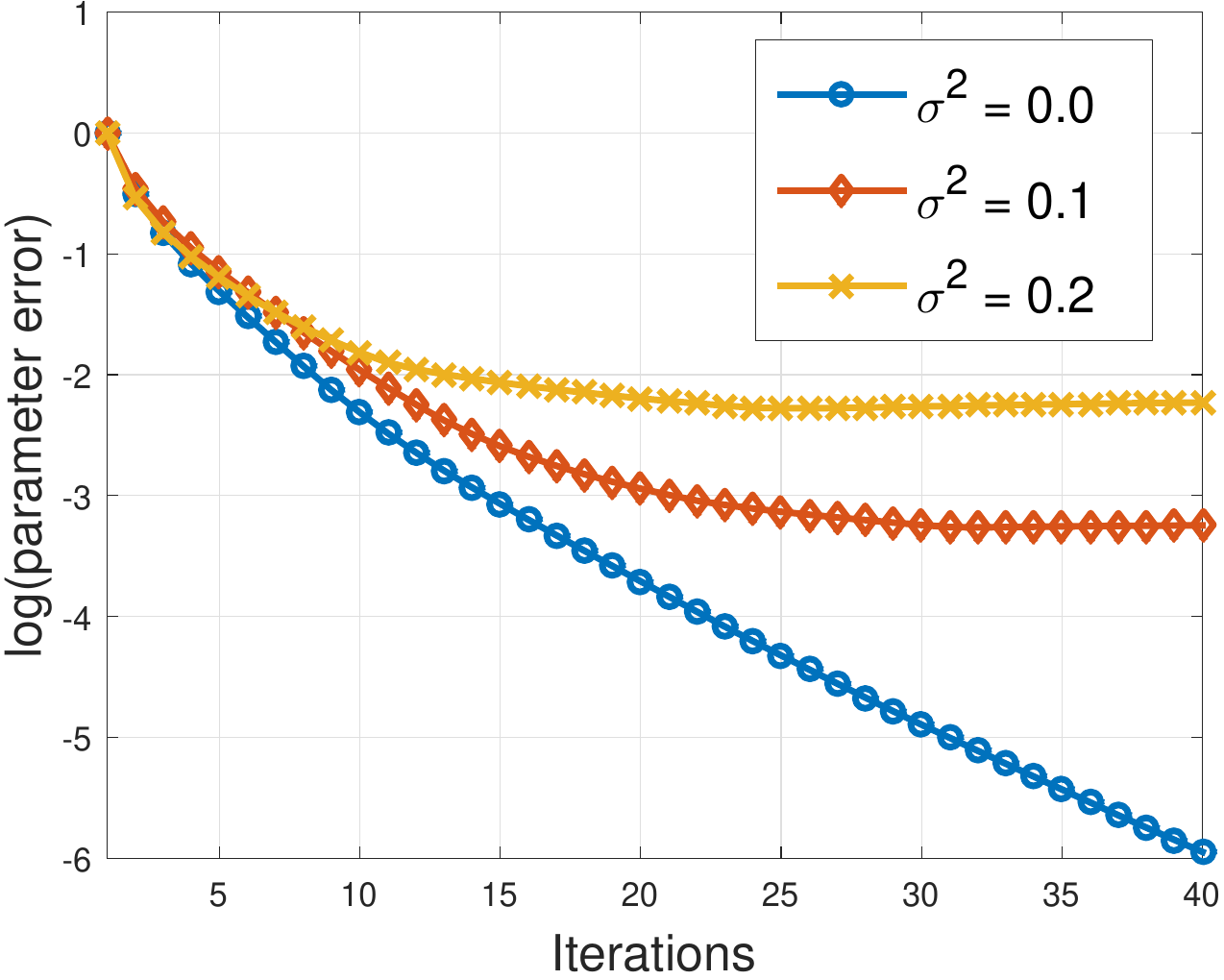}}\hspace{.1em}
\subfloat[][$\log(\norm{{\param^t} -\param^*}_2)$  vs. sample size.]{
\includegraphics[width=.47\columnwidth]{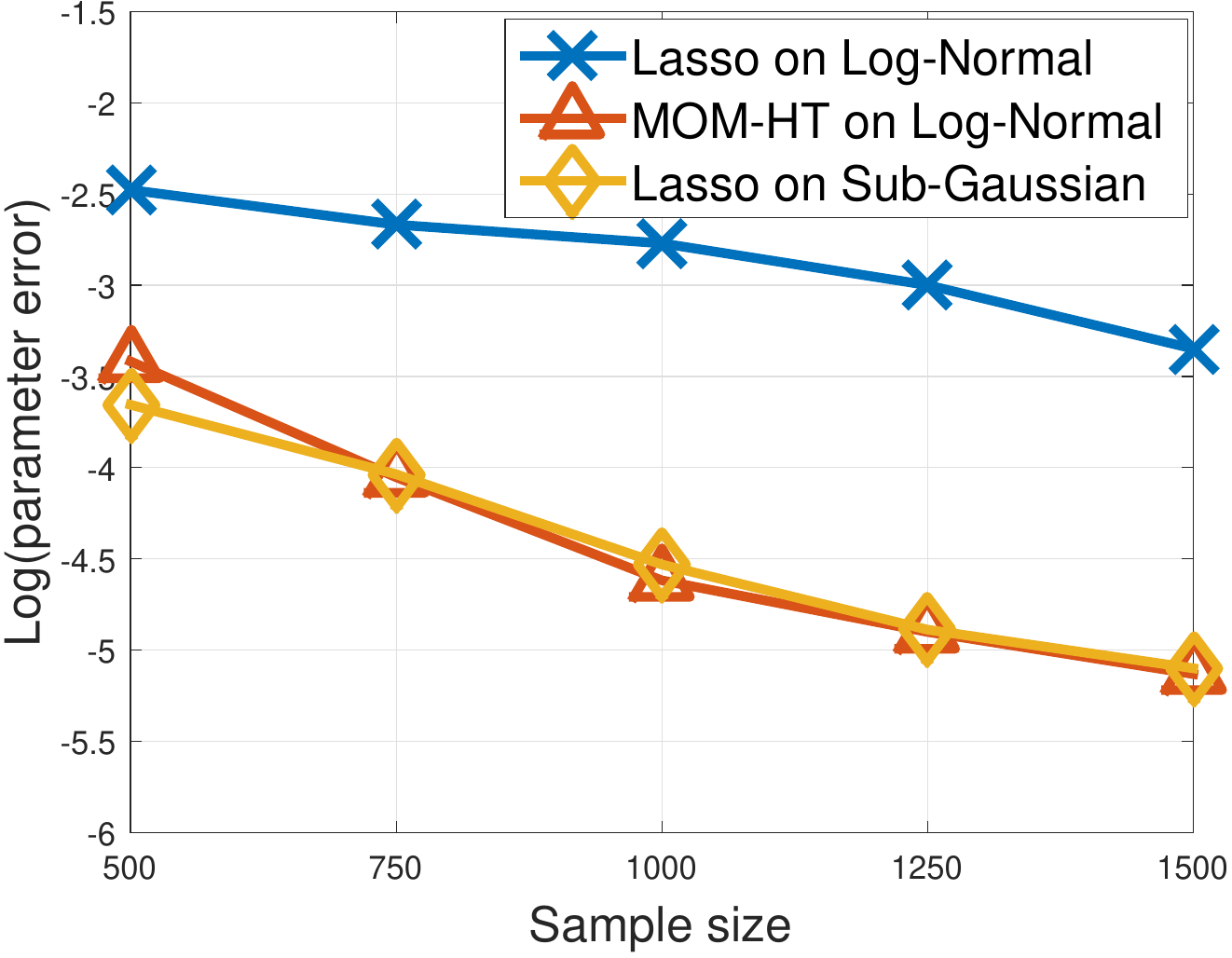}}
\caption{\footnotesize{
In the left plot, the corruption level $\eps$ is fixed and we use
\Cref{alg:THT} with trimming for different noise level $\sigma^2$.
In the right plot, we consider log-normal samples, and we use
\Cref{alg:THT} with MOM for different sample size to compare with baselines (Lasso on heavy tailed data, and Lasso on sub-Gaussian data).
}}
\label{fig:Linear}
\end{minipage} \hfill
\begin{minipage}{.48\textwidth}
  \centering
\subfloat[][Graph estimated by \Cref{alg:graphical_IHT}.]{
\includegraphics[width=.47\columnwidth]{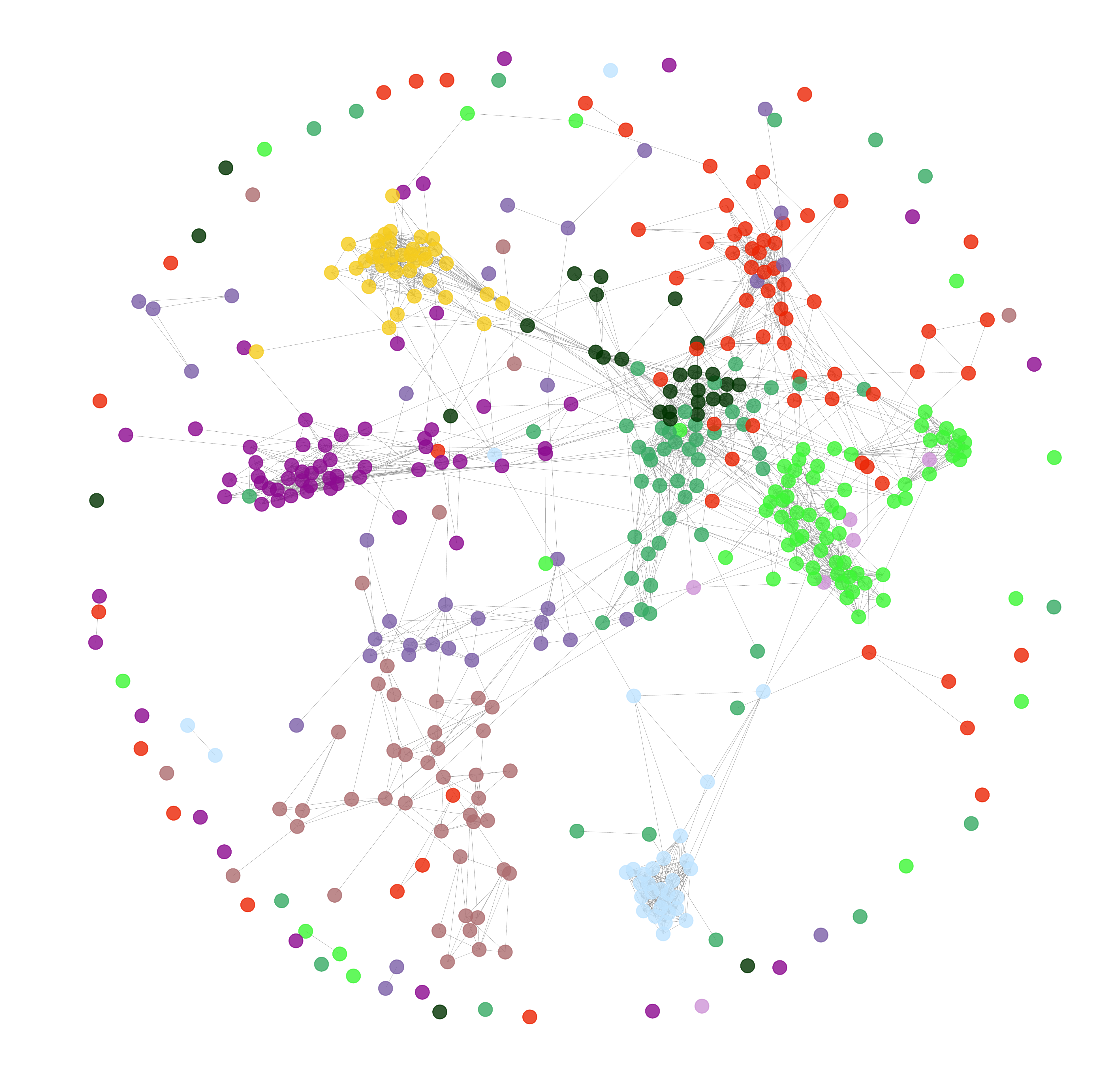}}\hspace{.1em}
\subfloat[][Graph estimated by Vanilla NS approach.]{
\includegraphics[width=.47\columnwidth]{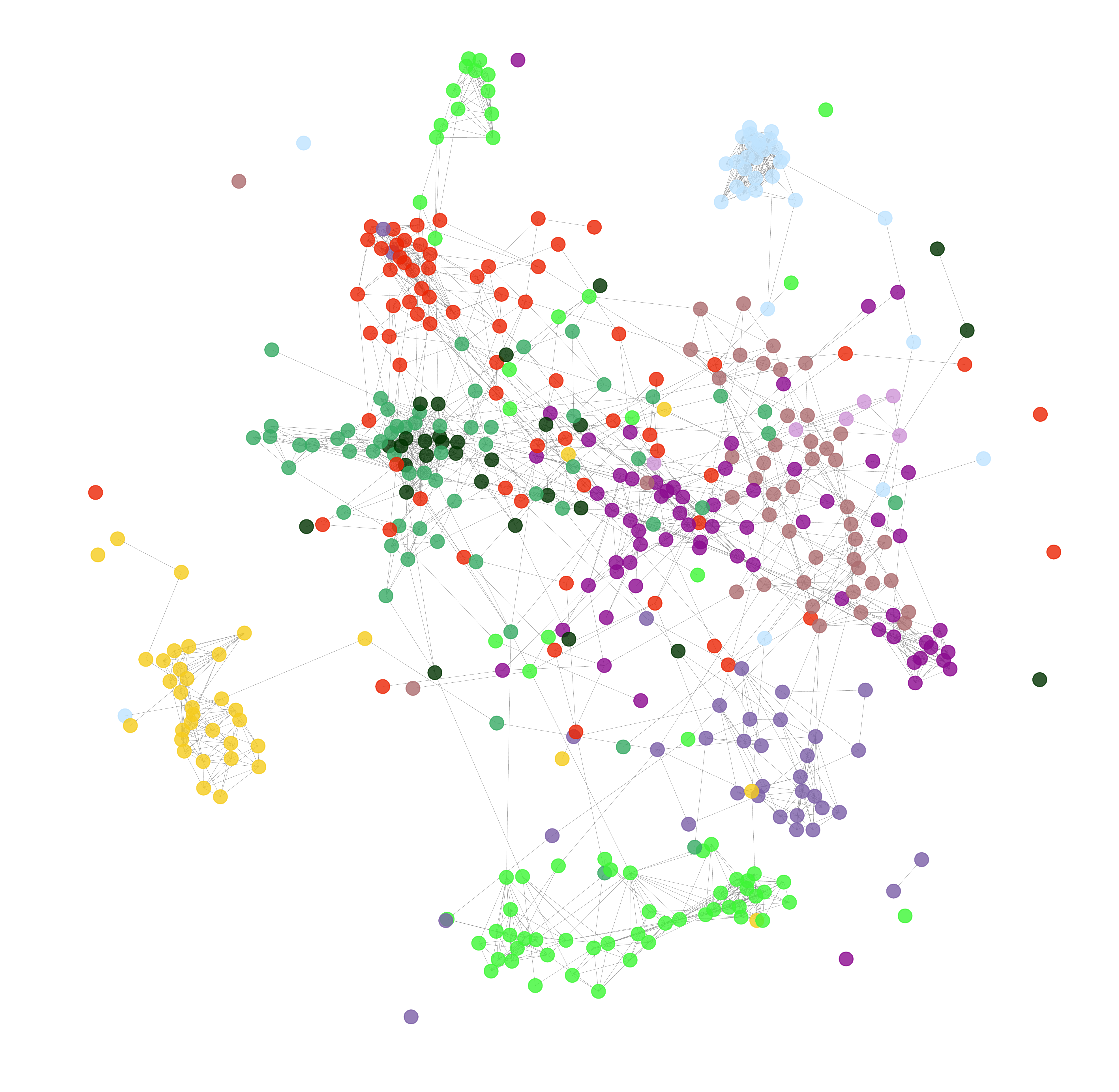}}
\caption{\footnotesize{
Graph estimated from the S\&P 500 stock data by \Cref{alg:graphical_IHT} and Vanilla NS approach. Variables are colored according to their sectors. In particular, the stocks from sector Information Technology are colored as purple.}}
\label{fig:additional_clustering}
\end{minipage}
\end{figure}

\textbf{Sparse regression with arbitrary corruption. }
We generate samples from 
a sparse  regression model (\Cref{model:linear}) with a Toeplitz covariance $\bm{\Sigma}$.
Here, the stochastic noise $\xi \sim \mathcal{N}(0, \sigma^2)$, and we vary the noise level $\sigma^2$ in different simulations.
We add outliers with $\eps=0.1$, and track the parameter error $\norm{{\param^t} -\param^*}_2$ in each iteration.
Left plot of \Cref{fig:Linear}
shows \Cref{alg:THT}'s linear convergence, and the error curves
flatten out at the final error level.         
Furthermore,  \Cref{alg:THT} can  achieve  machine precision when $\sigma^2 = 0$, which means exactly recovering of $\param^*$.

\textbf{Sparse regression with heavy tails. }
We consider a log-normal distribution (a typical example of heavy tails)  in \Cref{model:linear}. More specifically, 
$\bm{x}_i = \sqrt{\bm{\Sigma}} \widetilde{\bm{x}}_i$, and 
$\xi_i = \sigma \widetilde{\xi}_i$. Here,  $\Sigma$ is the same Toeplitz covariance, each entry of $\widetilde{\bm{x}}_i$ and 
$\widetilde{\xi}_i$ follows from $(Z - \Expe Z)/\sqrt{\Var(Z)}$, where $Z \sim \log\mathcal{N}(0, 4)$.
We fix $k, d, \sigma$, and vary sample size $n$.
For log-normal samples, we run \Cref{alg:THT} with MOM and vanilla Lasso. We then re-generate standard Gaussian samples using the same dimensions with $\bm{\Sigma}$
and run Vanilla Lasso.
Each curve in the right plot of \Cref{fig:Linear} is the average of 50 trials. 
\Cref{alg:THT} with MOM significantly improves vanilla Lasso on log-normal data, and has the same performance as Lasso on sub-Gaussian data

\textbf{Real data experiments. }
We next apply \Cref{alg:graphical_IHT}, to a US equities dataset \cite{equities2012nonparanormal, HugePackage}, which is heavy-tailed and has many outliers \cite{de2018advances}. 
The dataset contains 1,257  daily closing prices of 452 stocks (variables).
It is well known that stocks from the same sector  tend to be clustered together \cite{king1966market}.
Therefore, we use Robust NS (\Cref{alg:graphical_IHT}) to
construct an undirected graph among  stocks.
Graphs estimated by different algorithms are shown in \Cref{fig:additional_clustering}.
We can see that stocks from the same sector are clustered together, and these clustering centers can be easily identified. We also compare \Cref{alg:graphical_IHT} to the baseline NS approach (as in the ideal setting). We can observe that stocks from Information Technology (colored by purple) are much better clustered by  \Cref{alg:graphical_IHT}.

%% file: Proofs.tex
\setstretch{1.2}

\paragraph{Notations in Appendix.}    
In our proofs, the exponent $-10$ in tail bounds is arbitrary, and can be changed to other larger constant without affecting the results.   
 $\{c_j\}_{j=0}^3$  denote universal constants, and they may change line by line.

\input{MOM.tex}
\input{Concentration.tex}

\input{Convergence.tex}

\input{Appendix_Sparsity.tex}

%% file: MOM.tex
\section{Proofs for the gradient estimators}
\label{sec:proof_grad}

In Robust Hard Thresholding (\Cref{alg:THT}), we use  trimmed gradient estimator or MOM gradient estimator. And in \Cref{thm:meta}, the key quantity to control the statistical rates of convergence is the Robust Descent Condition 
(\Cref{def:RDC}). 

By Holder inequality, we have
\begin{align*}
\abs{ \inner{\RobustG(\param) - {\UMean}(\param)}{\widetilde{\param} - \param^*}} \leq \norm{{\RobustG(\param) - {\UMean}(\param)}}_\infty \norm{\widetilde{\param} - \param^*}_1.
\end{align*}
In this section, we provide one direct route for obtaining upper bound of Robust Descent Condition via bounding the infinity norm of the robust gradient estimate (\Cref{prop:contamination} and \Cref{prop:heavy}).

Later, in \Cref{sec:proof_stat}, we will leverage \Cref{prop:contamination} and \Cref{prop:heavy}
in verifying
the 
Robust Descent Condition for 
trimmed/MOM gradient estimator
under sparse linear/logistic regression.
Together with \Cref{thm:meta}, this will complete 
\Cref{thm:main_linear} -- \Cref{thm:main_logistic_heavy}.

\bpr\label{prop:contamination}
Suppose we observe $n$ $\eps$-corrupted sub-Gaussian samples (\Cref{def:contamination_model}).
With probability at least $1 - {d^{-3}}$, the coordinate-wise trimmed gradient estimator can
guarantee
\begin{itemize}[leftmargin=*]
  \item  $\norm*{\RobustG - \OGrad}_\infty = O\left( \sqrt{ \norm*{\param - \param^*}_2^2 + \sigma^2}
\left(
{ \eps \log (nd) } +
\sqrt{{\log d}/{n}} \right) \right)$  for
sparse linear regression (\Cref{model:linear}).
  \item $
\norm*{\RobustG - \OGrad}_\infty = O\left(
{ \eps \log (nd) } +
\sqrt{{\log d}/{n}} \right)$ for sparse logistic regression (\Cref{model:logistic}).
\end{itemize}
\epr


\bpr\label{prop:heavy}
Suppose we observe $n$ samples from the heavy tailed model with bounded 4-th moment covariates.
With probability at least $1 - {d^{-3}}$, the coordinate-wise Median of Means gradient estimator can
guarantee
\begin{itemize}[leftmargin=*]
  \item $\norm*{\RobustG - \OGrad}_\infty = O\left( \sqrt{ \condnum^2 \norm*{\param - \param^*}_2^2 + \condnum\sigma^2}
\sqrt{{\log d}/{n}} \right)$ for
sparse linear regression;
  \item  $
\norm*{\RobustG - \OGrad}_\infty = O\left( \sqrt{\condnum{\log d}/{n}} \right)$  for sparse logistic regression.
\end{itemize}
\epr

\subsection{Proofs for the MOM gradient estimator}
\label{sec:proof_MOM}

We first prove \Cref{prop:heavy}. \Cref{prop:contamination} of trimmed gradient estimator for $\eps$-corrupted sub-Gaussian samples has the same dependency on $\norm*{\param - \param^*}_2$. The proof of \Cref{prop:contamination} leverages standard  concentration bound for sub-Gaussian samples, and then uses trimming to control the effect of outliers.

\begin{proof}[Proof of \Cref{prop:heavy}]
For $\ell_2$ loss function,   we have $\gradsample(\param) = \x({\x}^\top{\param} - y)$, where we omit the subscript $i$ in the proof.
We denote $\Delta \coloneqq \param - \param^*$,
and bound the operator norm of the covariance of gradient samples
\begingroup
\allowdisplaybreaks
\begin{align*}
     \opnorm{{\Expe(\gradsample - \UMean) (\gradsample-\UMean)^\top}}
    &\leq  \opnorm{\Expe((\x\x^\top - \Sig) \Delta\Delta^\top (\x\x^\top - \Sig))} + \opnorm{ \Expe(\xi^2 \x\x^\top)}\\
    &\leq  \sup_{\vect_1 \in \mathcal{S}^{d-1}} \vect_1^\top {\Expe((\x\x^\top - \Sig) \Delta\Delta^\top (\x\x^\top - \Sig))} \vect_1 + \sigma^2 \opnorm{\Sig} \\
    &\leq  \sup_{\vect_1 \in \mathcal{S}^{d-1}}\inner{\Delta\Delta^\top}{\Expe (\x\x^\top - \Sig)\vect_1 \vect_1^\top (\x\x^\top - \Sig)}  + \sigma^2 \opnorm{\Sig} \\
    & \overset{(i)}\leq   \norm{\Delta}_2^2 \sup_{\vect_1, \vect_2 \in \mathcal{S}^{d-1}} {\Expe (\vect_2^\top (\x\x^\top - \Sig) \vect_1)^2}  + \sigma^2  \opnorm{\Sig} \\
    & \leq   2 \norm{\Delta}_2^2 \sup_{\vect_1, \vect_2 \in \mathcal{S}^{d-1}} ({\Expe (\vect_2^\top (\x\x^\top) \vect_1)^2} + \opnorm{\Sig}^2)  + \sigma^2  \opnorm{\Sig} \\
    &\leq   2  \norm{\Delta}_2^2 \sup_{\vect_1, \vect_2 \in \mathcal{S}^{d-1}} (\sqrt{\Expe (\vect_2^\top \x )^4 \Expe(\x^\top \vect_1)^4}
+ \opnorm{\Sig}^2 ) + \sigma^2  \opnorm{\Sig}\\
    & \overset{(ii)} \leq  2 (C_4+1) \opnorm{\Sig} ^2 \norm{ \Delta}_2^2+ \sigma^2  \opnorm{\Sig},
\end{align*}
\endgroup
where (i) follows from the Holder inequality, 
and (ii) follows from the 4-th moment bound assumption.

Hence, by using coordinate-wise  Median of Means gradient estimator,  we have 
\begin{align*}
\sup_{\vect \in \mathcal{S}^{d-1}} \vect^\top (\RobustG - \OGrad) \overset{(i)}{=} O\left( \sqrt{{ \opnorm{\Cov(\gradsample)} \log d}/{n}} \right)
= O\left(  \sqrt{ \condnum^2 \norm*{\param - \param^*}_2^2 + \condnum\sigma^2}
\sqrt{{\log d}/{n}} \right)
\end{align*}
with probability at least $1 - {d^{-4}}$,
where (i) follows from Proposition 5  in 
\cite{hsu2016loss}.
Applying union bounds on all d indexes, we have
$\norm*{\RobustG - \OGrad}_\infty = O\left(  \sqrt{ \condnum^2 \norm*{\param - \param^*}_2^2 + \condnum\sigma^2}
\sqrt{{\log d}/{n}} \right)$ with probability at least $1 - {d^{-3}}$.

For logistic loss,
 the gradient can be computed as:
$\gradsample = \frac{- y \bm{x} }{1 + \exp\left(y\bm{x}^{\top} \param \right)},
$
where we omit the subscript $i$
in the proof.

Since $y \in \{-1, +1\}$, and ${1 + \exp\left(y\bm{x}^{\top} \param \right)} \geq 1$,
we directly have
$\norm*{{\Expe(\gradsample - \UMean) (\gradsample-\UMean)^\top}}_{\rm op} \leq  \opnorm{\Sig}.$
Similar to the case of $\ell_2$ loss, we have
$\norm*{\RobustG - \OGrad}_\infty = O\left( \sqrt{\condnum {\log d}/{n}} \right)$,
with probability at least $1 - {d^{-3}}$.

\end{proof} 

%% file: Concentration.tex
\subsection{Proofs for the trimmed gradient estimator}
\label{sec:proof_trim}

We then turn to the trimmed gradient estimator for the case of arbitrary corruption.
Before we proceed to the trimmed estimator, 
let us  first visit the definition and tail bounds of sub-exponential random variable,
as it will be used in sparse linear regression, where the gradient's distribution is indeed sub-exponential under the sub-Gaussian  assumptions in
\Cref{model:linear}.

We first present standard concentration inequalities
    (\cite{WainwrightBook}).
\bde[Sub-exponential random variables]
A random variable $X$ with mean $\mu$ is sub-exponential if there
are non-negative parameters $\nu$ such that
\begin{align*}
  \Expe[\exp \left(t\left(X - \mu\right) \right)] \leq \exp\left(\frac{\nu^2 t^2}{2}\right), \quad \text{for all } \abs{t} < \frac{1}{\nu}.
\end{align*}
\ede

\ble[Bernstein's inequality]
\label{lemma:Bernstein}
Suppose that $X_i, i=1,\cdots,n$,  are i.i.d. sub-exponential random variables with parameters $\nu$. Then
\begin{subequations}
\begin{empheq}[left={\Pr\left( \frac{1}{n}\sum_{i=1}^{n}X_i \geq \mu + t\right)\leq \empheqlbrace\,}]{align}
     & \exp\left(-\frac{nt^2}{2\nu^2}\right) \quad  \text{if } 0\leq t\leq \nu, \text{ and} \nonumber\\
     & \exp\left(-\frac{nt}{2\nu}\right) \quad  \text{ for }  t> \nu. \nonumber
\end{empheq}
\end{subequations}
We also have a two-sided
tail bound
\begin{align*}
  \Pr\left( \abs{\frac{1}{n}\sum_{i=1}^{n}X_i - \mu} \geq t\right)\leq 2 \exp\left( -n \min\left(\frac{t^2}{2\nu^2}, \frac{t}{2\nu}\right)\right).
\end{align*}
\ele

We define $\alpha$-trimmed mean estimator for one dimensional samples, and denote it as $\trim_\alpha(\cdot)$.
\bde[$\alpha$-trimmed mean estimator]
\label{def:trim_mean}
Given  a set of $\eps$-corrupted samples $\{\bm{z}_i \in \Real, i\in\Input\}$, the
coordinate-wise trimmed mean estimator $\trim_\alpha(\cdot)$
removes the
largest and smallest $\alpha$ fraction of elements in
$\{\bm{z}_i \in \Real, i\in\Input\}$,
and calculate the mean of the remaining terms.
We choose $\alpha = c_0\eps$,
for a constant $c_0\geq 1$.
We also require that
$\alpha \leq 1/2 - c_1$, for some small constant $c_1 > 0$.
\ede

\Cref{prop:trimmed} shows the  guarantees for this robust gradient estimator in each coordinate.
We note that   
 \Cref{prop:trimmed} is stronger than  guarantees for trimmed mean estimator (Lemma 3) in \cite{Yin_median}.
 In our contamination model \Cref{def:contamination_model},
the adversary may \emph{delete} $\eps$-fraction of authentic samples, and then add arbitrary outliers.
 And \Cref{prop:trimmed} provides   guarantees for trimmed mean estimator on sub-exponential random variables.    
The trimmed mean estimator is robust enough, that it allows the adversary to arbitrarily remove $\epsilon$-fraction of data points.
We use $\OGood^j$ to denote the $\Real^1$ samples at the $j$-th coordinate of $\OGood$.
We can also define $\Input^j$ in the same way.

\ble\label{prop:trimmed}
Suppose we observe $n = \Omega(\log d)$  $\eps$-corrupted samples from \Cref{def:contamination_model}.
For each dimension
 $j \in \{1, 2, \cdots, d\}  $, we assume
the samples  in
 $\OGood^j$ are i.i.d. $\nu$-sub-exponential with mean $\bm{\mu}^j$.
 After the contamination, we have
the $j$-th $\Real^1$ samples as $\Input^j$.
Then, we can guarantee the trimmed mean estimator on $j$-th dimension that
\begin{align*}
\left|{ \trim_\alpha\{{x}_i : i\in \Input^j\} - \bm{\mu}^j  } \right|  =O\left(\nu \left( \eps \log (nd) + \sqrt{\frac{\log d}{n}} \right) \right)
\end{align*}
with probability at least $1 - {d^{-4}}$.
\ele

We leave the proof of
\Cref{prop:trimmed} at the end of this section.
Then, we present analysis of trimmed gradient estimator for sparse linear regression and sparse logistic regression
by using \Cref{prop:trimmed}.
For
 sparse linear regression
model with sub-Gaussian covariates,
the distribution of authentic gradients are sub-exponential instead of sub-Gaussian.
More specifically, we first prove that when the current parameter iterate is $\param$, the  sub-exponential parameter of all authentic gradient is
$O(( \norm{\Delta}_2^2 + \sigma^2)^{1/2})$, where
$\Delta \coloneqq \param - \param^*$.

To gain some intuition for this, we can consider the sparse linear equation problem, where $\sigma^2 = 0$.
When $\param = \param^* ( \norm{\Delta}_2^2 = 0)$, we exactly recover $\param^*$, and all stochastic gradients
of authentic samples are actually zero vectors,
as all observations are noiseless.
It is clear that we will have
sub-exponential parameter as $0$.


\spro[Proof of \Cref{prop:contamination}]
For any $\param$, the gradient for one sample can be written as
$$
\gradsample =  \bm{x}\left(\bm{x}^{\top} \param - y\right), \text{ and } \UMean = \Expe(\gradsample) =  \bm{\Sigma} \left(\param - \param^*\right),
$$
where we omit the subscript $i$
in the proof. For any
fixed standard basis vector
$\vect \in\mathbb{S}^{d-1}$, and define
$\Delta = \param - \param^*$, we have
\begin{align}
  \vect^\top\gradsample = \vect^\top \bm{x}\bm{x}^{\top} \Delta - \vect^\top\bm{x}\xi,\text{ and } \vect^\top \UMean =  \vect^\top \bm{\Sigma} \Delta.
\end{align}
To characterize the tail bounds of $\vect^\top\gradsample$, we study the moment generating function:
\begin{align*}
    \Expe[\exp\left( t \left(\vect^\top\gradsample - \vect^\top \UMean\right)\right)] = \Expe[\exp\left( t \left(\vect^\top \left(\bm{x}\bm{x}^{\top} - \bm{\Sigma}\right) \Delta - \vect^\top\bm{x}\xi\right)\right)].
\end{align*}
We denote $\gamma \in \{-1, +1\}$ as a Rademacher random variable, which is independent of $\bm{x}$ and $\xi$. Then we can use a standard symmetrization technique \cite{WainwrightBook},
 \begin{align*}
 \mathbb{E}_{\bm{x}, \xi}[\exp\left( t \left(\vect^\top \left(\bm{x}\bm{x}^{\top} - \bm{\Sigma}\right) \Delta - \vect^\top\bm{x}\xi\right)\right)] &\leq
  \mathbb{E}_{\bm{x}, \xi, \gamma}[\exp\left( 2t \gamma\left(\vect^\top\bm{x}\bm{x}^{\top} \Delta - \vect^\top\bm{x}\xi\right)\right)]\\
  & \overset{(i)}{=} \sum_{k = 0}^{\infty}
  \frac{1}{k!}\left(2t\right)^k \Expe[\gamma^k \left(\vect^\top\bm{x}\bm{x}^{\top} \Delta - \vect^\top\bm{x}\xi\right)^k] \\
  & \overset{(ii)}{=} 1 + \sum_{l=1}^{\infty}\frac{1}{\left(2l\right)!}\left(2t\right)^{2l} \Expe[ \left(\vect^\top\bm{x}\right)^{2l} \left(\bm{x}^{\top} \Delta - \xi\right)^{2l}],
\end{align*}
where $(i)$ follows from the exponential function's power series expansion, and
$(ii)$ follows from the independence of $\gamma$, together with the fact that all odd moments of the $\gamma$ terms have zeros means.

By the Cauchy-Schwarz inequality, we have
\begin{align*}
    \Expe[ \left(\vect^\top\bm{x}\right)^{2l} \left(\bm{x}^{\top} \Delta - \xi\right)^{2l}] \leq
    \sqrt{\Expe[ \left(\vect^\top\bm{x}\right)^{4l} ] \Expe[  \left(\bm{x}^{\top} \Delta - \xi\right)^{4l}]}.
\end{align*}

It is clear that $\xi$ is
a sub-Gaussian random variable with
parameter $\sigma$.
Since $\bm{x} \sim \mathcal{N} \left(0, \bm{\Sigma} \right)$, we have
$\vect^\top\bm{x} \sim \mathcal{N} \left(0, \vect^\top\bm{\Sigma} \vect \right)$.
For
any
fixed standard basis vector
$\vect \in\mathbb{S}^{d-1}$, we can conclude that $\vect^\top\bm{x}$ is sub-Gaussian with parameter at most $1$ based on
\Cref{model:linear}.
By basic properties of
sub-Gaussian random variables
\cite{WainwrightBook}, we have
\begin{align*}
    \sqrt{\Expe[ \left(\vect^\top\bm{x}\right)^{4l} ] } &\leq
    \sqrt{
    \frac{\left(4l\right)!}{2^{2l} \left(2l\right)!} }\left(\sqrt{8}e\right)^{2l} \\
    \sqrt{\Expe[  \left(\bm{x}^{\top} \Delta - \xi\right)^{4l}]} &\overset{(i)}{\leq}
    \sqrt{
    \frac{\left(4l\right)!}{2^{2l} \left(2l\right)!} }\left(8e^2\left( \norm{\Delta}_2^2 + \sigma^2\right)\right)^l,
\end{align*}
where $(i)$ follows from the fact that $\bm{x}^{\top} \Delta - \xi$ is the weighted summation of two independent sub-Gaussian random variables. Hence, we have
\begin{align}
    \Expe[\exp\left( t \left(\vect^\top\gradsample - \vect^\top \UMean\right)\right)] &\leq 1 + \sum_{l=1}^{\infty}\frac{1}{\left(2l\right)!}\left(2t\right)^{2l} \frac{\left(4l\right)!}{2^{2l} \left(2l\right)!}\left(\sqrt{8}e\right)^{4l}
    \left( \norm{\Delta}_2^2 + \sigma^2\right)^l \nonumber\\
    & \overset{(i)}{\leq} 1 + \sum_{l=1}^{\infty}
    \left(4t\right)^{2l}\left(\sqrt{8}e\right)^{4l}
    \left( \norm{\Delta}_2^2 + \sigma^2\right)^l \nonumber\\
    & =  1 + \sum_{l=1}^{\infty}
    \left(4t\right)^{2l}\left({8}e^2\right)^{2l}
    \left(\sqrt{ \norm{\Delta}_2^2 + \sigma^2}\right)^{2l}, \label{equ:geo_sum}
\end{align}
where $(i)$ follows from
$\left(4l\right)! \leq 2^{4l}\left(\left(2l\right)!\right)^2$ (proof by mathematical induction).
When we have 
$f\left(t\right) = {32 t e^2 \sqrt{ \norm{\Delta}_2^2 + \sigma^2}}  < 1$,
\cref{equ:geo_sum} converges to$\frac{1}{1 - f^2\left(t\right)}$.
Hence,
\begin{align*}
    \Expe[\exp\left( t \left(\vect^\top\gradsample - \vect^\top \UMean\right)\right)] \leq \frac{1}{1 - f^2\left(t\right)} \leq \exp\left(f^2\left(t\right)\right).
\end{align*}
That being said, $ \vect^\top\gradsample$ is a sub-exponential random variable.
By choosing $\vect$ as each coordinate in $\Real^d$, each coordinate of gradient has sub-exponential parameter as $ 32 \sqrt{2}  e^2 \sqrt{ \norm{\Delta}_2^2 + \sigma^2}$.

Then, applying \Cref{prop:trimmed} on this collection of
corrupted
sub-exponential random variables,  we have
\begin{align}
\left|{ \trim_\alpha\{{x}_i : i\in \Input^j\} - \bm{\mu}^j  } \right| = O\left(\sqrt{ \norm{\Delta}_2^2 + \sigma^2}
\left( \eps \log (nd) + \sqrt{\frac{\log d}{n}} \right) \right),
\label{equ:linear_ineq_1}
\end{align}
with probability at least $1-{d^{-4}}$.


Applying union bounds on  \cref{equ:linear_ineq_1} for all $d$ indexes, we have
\begin{align*}
\norm{\RobustG - \OGrad}_\infty = O\left(\sqrt{ \norm{\Delta}_2^2 + \sigma^2}
\left( \eps \log (nd) + \sqrt{\frac{\log d}{n}} \right)  \right),
\end{align*}
with probability at least $1-{d^{-3}}$.


In this subsection, we use
\Cref{prop:trimmed} to
bound $\norm*{\RobustG - \OGrad}_\infty$ for sparse logistic regression. The technique for sparse logistic regression is similar to linear regression. Since we can directly show the sub-Gaussian distribution of gradient in this case,
applying \Cref{prop:trimmed} leads to the bound for $\norm*{\RobustG - \OGrad}_\infty$.



Under the statistical model of sparse logistic regression, the gradient can be computed as:
\begin{align*}
\gradsample = \frac{- y \bm{x} }{1 + \exp\left(y\bm{x}^{\top} \param \right)},
\end{align*}
where we omit the subscript $i$
in the proof.

Since $y \in \{-1, +1\}$, and ${1 + \exp\left(y\bm{x}^{\top} \param \right)} \geq 1$,
then for any
fixed standard basis vector
$\vect \in\mathbb{S}^{d-1}$,  $\vect^\top\gradsample$
 is sub-Gaussian with parameter at most $1$
 based on \Cref{model:logistic}. Notice that $\nu$-sub-Gaussian random variables are
still $\nu$-sub-exponential.
 Applying \Cref{prop:trimmed} again, we have
\begin{align}
\left|{ \trim_\alpha\{{x}_i : i\in \Input^j\} - \bm{\mu}  } \right| = O
\left( \eps \log (nd) + \sqrt{\frac{\log d}{n}} \right)
\label{equ:logistic_ineq_1}
\end{align}
with probability at least $1 - {d^{-4}}$.

Applying union bounds on \cref{equ:logistic_ineq_1}  for all $d$ indexes, we have
\begin{align*}
\norm{\RobustG - \OGrad}_\infty = O
\left( \eps \log (nd) + \sqrt{\frac{\log d}{n}} \right)
\end{align*}
with probability at least $1-{d^{-3}}$.
\fpro

\subsection{Trimmed mean estimator for strong contamination model}

Now, it only remains to prove \Cref{prop:trimmed}.
The proof technique is as follow: even though an adversary may delete samples from $\OGood^j$, we can still show the concentration inequalities for remaining authentic $\Real^1$ samples (denoting as $\Good^j$ in the proof).
Then, we show that by using trimmed mean estimator, either the abnormal outliers will be removed, or their effect is controlled.

\spro[Proof of \Cref{prop:trimmed}]
Without loss of generality, we assume
$\bm{\mu} = 0$ throughout the proof.

 For each dimension $j \in \{1, 2, \cdots, d\}$,
we can split the $j$-th one-dimensional samples as
$\Input^j = \Good^j \bigcup \Bad^j$.
To study the performance of $\trim_\alpha\{{x}_i : i\in \Input^j\}$,
we first show a
concentration inequality of the sub-exponential variables in $\Good^j$, without
worrying about removing points from $\OGood^j $.
This part of our proof is similar to Lemma 4.5 in \cite{Moitra2016FOCS}.

\paragraph{Concentration inequality for $\Good^j$}

We consider the set $\{{x}_i : i\in \OGood^j \}$ in $\Real^1$.
Since $\Good^j$ is a subset of $\OGood^j$, by triangle inequality we have,
\begin{align*}
   \abs{{\Expe_{i \in_u \Good^j} x_i}}
    = \abs{\frac{\sum_{i \in \Good^j} x_i}{\left(1-\eps\right)n}}
    \leq \underbrace{\abs{\frac{\sum_{i \in \OGood^j} x_i }{\left(1-\eps\right)n} }}_{A_1} + \underbrace{\abs{\frac{\sum_{i \in \OGood^j \setminus \Good^j} x_i }{\left(1-\eps\right)n} }}_{A_2}.
\end{align*}
The first term $A_1$ is simply
the average of i.i.d.
sub-exponential random variables. By \Cref{lemma:Bernstein}, we have
\begin{align}
\Pr\left(
 \abs{{\frac{\sum_{i \in \Good^j} x_i}{\left(1-\eps\right)n}}} \geq { c_0 \nu}
 \sqrt{\frac{\log d}{n}} \right) &\leq  2 \exp\left(-c_1 n \min \left(\sqrt{\frac{\log d}{n}}, {\frac{\log d}{n}} \right) \right) \\
 &\leq {c_2}{d^{-10}}.
 \label{equ:A_1}
\end{align}

For the second term $A_2$,
We now wish to show that with probability $1-\tau$, there does not exist a subset $\Good^j$ so that the $A_2$ is
more than $ \delta_0 $. This event is equivalent to
 \begin{align*}
   \abs{\frac{\sum_{i \in \OGood^j \setminus \Good^j} x_i }{\left(1-\eps\right)n} } =  \abs{\frac{\sum_{i \in \OGood^j \setminus \Good^j} x_i }{\eps n} } \frac{\eps}{1 - \eps} \geq \delta_0
 \end{align*}

Let $\delta_1 = \frac{1 - \eps}{\eps} \delta_0$. For one subset $\OGood^j \setminus \Good^j$, by \Cref{lemma:Bernstein}, we have
\begin{align*}
\Pr\left(
  \abs{\frac{\sum_{i \in \OGood^j \setminus \Good^j} x_i }{\eps n} } \geq  \delta_1 \right) \leq 2 \exp\left(- \eps  n \min\left(\frac{\delta_1^2}{2\nu^2}, \frac{\delta_1}{2\nu}\right)\right).
\end{align*}
Then, we take union bounds over all possible $\OGood^j \setminus \Good^j$,  which have $\binom{n}{\epsilon n}$ events.
Hence, the tail probability of $A_2$ can be bounded as
\begin{align}
\tau &\leq 2
\binom{n}{\epsilon n} \exp\left(-\eps  n \min\left(\frac{\delta_1^2}{2\nu^2}, \frac{\delta_1}{2\nu}\right)\right) \nonumber\\
&\overset{(i)}{\leq} c_0  \exp\left( n H(\eps) - \eps  n \min\left(\frac{\delta_1^2}{2\nu^2}, \frac{\delta_1}{2\nu}\right)\right)
\label{equ:A_2}
\end{align}
where (i) follows from the fact that $ \log \binom{n}{\epsilon n} = O (n H(\eps))$ for $n$ large enough, and $H(\cdot)$ is the binary entropy function. Choosing $\delta_1 = c_1 \nu \log (nd)$, and hence $\delta_0 = c_1 \nu \eps \log (nd)$,
we have $\tau \leq c_0 \exp (-c_2 n \eps \log (nd)) \leq c_3 {d^{-10}}$.

Combining the analysis on $A_1$ and $A_2$ (\cref{equ:A_1} and \cref{equ:A_2}), we have
\begin{align}
\Pr\left(
 \abs{{\Expe_{i \in_u \Good^j} x_i}} \geq \nu \left( c_0
 \sqrt{\frac{\log d}{n}} + c_1 \eps \log (nd) \right) \right) \leq {c_2}{d^{-10}}.
 \label{equ:B_1}
\end{align}
This completes the concentration bounds on $\abs{{\Expe_{i \in_u \Good^j} x_i}}$ for all possible samples in $\Good^j$ without worrying about sample removing.

\paragraph{Trimmed mean estimator for $\Input^j$}

Then, we can consider the  contribution of each part in  $\Input^j = \Good^j \bigcup \Bad^j$.
We denote the remaining set after trimming as $\Remain^j$, and the trimmed set as $\Trim^j$.
Recall that we assume
$\bm{\mu} = 0$, we only need to bound $\abs{{ \trim_\alpha\{{x}_i : i\in \Input^j\}}}$, which is the empirical average of all samples in the remaining set
$\{{x}_i : i\in \Remain^j\}$.

As $\Remain^j$ can be easily separated by the union of two distinct set
$\Bad^j \bigcap \Remain^j$ and $\Good^j \bigcap \Remain^j$, we have the following inequalities,

\begin{align*}
\abs{{ \trim_\alpha\{{x}_i : i\in \Input^j\}}}  &= \abs{\frac{1}{ (1-2\alpha) n} \sum_{i \in \Remain^j} {x}_i} \\
&\leq\frac{1}{ (1-2\alpha) n} \abs{ \sum_{i \in \Good^j} {x}_i - \sum_{i \in \Good^j \bigcap \Trim^j} {x}_i + \sum_{i \in \Bad^j \bigcap \Remain^j} {x}_i }\\
&\leq\frac{1}{ (1-2\alpha) n} \left(  \underbrace{\abs{ \sum_{i \in \Good^j} {x}_i}}_{B_1} + \underbrace{\abs{ \sum_{i \in \Good^j \bigcap \Trim^j} {x}_i}}_{B_2}  + \underbrace{\abs{\sum_{i \in \Bad^j \bigcap \Remain^j} {x}_i }}_{B_3}\right)
\end{align*}

For any $i\in \Good^j$,
by \Cref{lemma:Bernstein}, we have
\begin{align*}
\Pr\left(
 \abs{x_i} \geq { c_0 \nu \log (nd)}
 \right) &\leq 2 \exp\left(-c_1 \min \left({\log (nd)}, {\log^2 (nd)} \right) \right).
\end{align*}
Applying a union bound for all samples,  we can control the maximum magnitude for any $i\in \Good^j$,
\begin{align*}
\Pr\left(
 \max_{i\in \Good^j} \abs{x_i} \geq { c_0 \nu \log (nd)}
 \right) &\leq 2 (1-\eps) n \exp\left(-c_1 \min \left({\log (nd)}, {\log^2 (nd)} \right) \right), \\
 &\leq c_1{d^{-10}}.
\end{align*}

We can bound $B_1$ by applying \cref{equ:B_1}. For the trimmed good samples $\{i \in \Good^j \bigcap \Trim^j\}$, we have $B_2 \leq 2 \alpha n \max_{i\in \Good^j} \abs{x_i} $. Since we choose $\alpha \geq \eps$, we have
$B_3 \leq \eps n \max_{i\in \Good^j} \abs{x_i}$.

Putting together the pieces,
and choosing $\alpha = c\eps$ for some universal constant $c\geq 1$,  we have
\begin{align*}
\left|{ \trim_\alpha\{{x}_i : i\in \Input^j \} - \bm{\mu}^j  } \right|  = O\left(\nu \left( \eps \log (nd) + \sqrt{\frac{\log d}{n}} \right) \right),
\end{align*}
with probability at least
$1 - d^{-4}$.
This completes the proof for
\Cref{prop:trimmed}.
\fpro

%% file: Convergence.tex
\section{Statistical estimation
via Robust Hard Thresholding
}
\label{sec:proof_stat}

Here,
we provide the Meta-Theorem \Cref{thm:meta} for statistical estimation performance of \Cref{alg:THT} under statistical models.

We first introduce a supporting Lemma on the property
of hard thresholding operator.

\ble[Lemma 1 in \cite{barber2018between}]
\label{lem:Hard}
We set $k'$ in hard thresholding operator as  $k' = k\slack^2$, where $\slack \geq 1$, then we have
\begin{align*}
 \sup\left\{\frac{\inner{\parambestk-\thr\left(\z\right)}{\z-\thr\left(\z\right)}}{\norm{\parambestk-\thr\left(\z\right)}^2_2} \ : \ \parambestk,\z\in\Real^d, \  \norm{\parambestk}_0\leq k, \ \parambestk\neq \thr\left(\z\right)\right\} = \frac{1}{2} \sqrt{\frac{k}{k'}} = \frac{1}{2 \slack}.
\end{align*}

\ele

Note that $\slack$  will be specified as $2 \condnum$ later in the proof,
as we choose $k' = 4\condnum^2 k$ as in \Cref{def:population risk}.





\spro[Proof of \Cref{thm:meta}]

We first study the objective function gap
$\obj\left(\parambar^t\right) - \obj\left(\parambestk\right)$.
Since the population risk $\obj$ satisfies $\sconvexity$-strong convexity and $\ssmoothness$-smoothness (\Cref{def:population risk}), we have
\begin{align}
\obj\left(\parambestk\right)&\overset{(i)}{\geq} \obj\left(\param^{t-1}\right)+\langle \OGrad\left(\param^{t-1}\right),\parambestk-\param^{t-1}\rangle+\frac{\sconvexity}{2}\norm{\parambestk-\param^{t-1}}^2_2,
\label{equ:ineq_1}\\
\obj\left(\parambar^t\right)&\leq \obj\left(\param^{t-1}\right)+\langle \OGrad\left(\param^{t-1}\right),\parambar^t-\param^{t-1}\rangle+\frac{\ssmoothness}{2}\norm{\parambar^t-\param^{t-1}}^2_2,
\label{equ:ineq_2}
\end{align}
where (i) follows from the fact that $\parambestk, \param^{t-1} \in \constraint$, and $\sconvexity$-strong convexity holds.

Combining these two inequalities, we obtain
\begin{align*}
  \obj\left(\parambar^t\right) - \obj\left(\parambestk\right)
 \leq \langle \OGrad\left(\param^{t-1}\right), \parambar^t - \parambestk\rangle -  \frac{\sconvexity}{2}\norm{\parambestk-\param^{t-1}}^2_2 +\frac{\ssmoothness}{2}\norm{\parambar^t-\param^{t-1}}^2_2.
\end{align*}

Expanding the last term,
we also have
\begin{align*}
\frac{\ssmoothness}{2} \norm{\parambar^t - \parambestk}^2_2
&= \frac{\ssmoothness}{2}
\norm{\param^{t-1} - \parambestk}^2_2 -  \frac{\ssmoothness}{2}  \norm{\parambar^t - \param^{t-1}}^2_2  + \ssmoothness \langle  \param^{t-1} -\parambar^t , \parambestk -\parambar^t\rangle\\
&= \frac{\ssmoothness}{2}
\norm{\param^{t-1} - \parambestk}^2_2 -  \frac{\ssmoothness}{2}  \norm{\parambar^t - \param^{t-1}}^2_2 \\
& \quad +\ssmoothness \underbrace{  \left\langle \left(\param^{t-1} - \eta \RobustG\left(\param^{t-1}\right)\right) - \parambar^t, \parambestk- \parambar^t\right\rangle}_{T_1}
-   \underbrace{ \left
\langle \RobustG\left(\param^{t-1}\right),\parambar^t-\parambestk \right \rangle}_{T_2}
\end{align*}

For the  term $T_1$,
recall that $\parambar^t$ is obtained from hard thresholding,  and
$\parambar^t = \thr\left(\param^{t-1}-\eta
\RobustG\left(\param^{t-1}\right)\right)$,
we apply \Cref{lem:Hard} with
$\z = \param^{t-1} - \eta \RobustG\left(\param^{t-1}\right)$:
\begin{align*}
   \left\langle \left(\param^{t-1} - \eta \RobustG\left(\param^{t-1}\right)\right) - \parambar^t, \parambestk- \parambar^t\right\rangle
    \leq \frac{1}{2\slack} \norm{\parambar^t - \parambestk}^2_2.
\end{align*}

The term $T_2$ can be bounded
by using \cref{equ:RDC} in \Cref{def:RDC}. We have
 \begin{align*}
\langle \RobustG\left(\param^{t-1}\right),\parambar^t-\parambestk\rangle &\geq \left\langle \OGrad\left(\param^{t-1}\right),\parambar^t-\parambestk \right\rangle -  \left|
\langle{\OGrad}\left(\param^{t-1}\right) - \RobustG\left(\param^{t-1}\right),  \parambar^t-\parambestk \rangle \right| \\
&\geq \left\langle \OGrad\left(\param^{t-1}\right),\parambar^t-\parambestk  \right \rangle -
    \left( \alpha \norm{\param^{t-1} - \param^*}_2 + \psi \right) \norm{\parambar^t- \param^*}_2,
\end{align*}
with probability at least $1-d^{-3}$.

We denote $\Deltabar^t = \parambar^t-\parambestk$ and
$\Delta^t = \param^t-\parambestk$.
Since, $\eta\sconvexity\geq \frac{1}{\ssmoothness}\cdot \sconvexity = \frac{1}{\condnum}$,
putting together the pieces,
we have
\begin{align}
\label{equ:iterates}
   \obj\left(\parambar^t\right) - \obj\left(\parambestk\right) \leq  \frac{\ssmoothness}{2}\left[\left(1-\frac{1}{\condnum}\right)\norm{\Delta^{t-1}}^2_2 -
\left(1 - \frac{1}{\slack} \right)\norm{\Deltabar^t}^2_2 \right]   +
\left( \alpha \norm{\Delta^{t-1}}_2 + \psi \right) \norm{\Deltabar^t}_2,
\end{align}
with probability at least $1-d^{-3}$.
Applying convexity, $\obj\left(\parambar^t\right) - \obj\left(\parambestk\right) \geq 0$, as $\parambestk$ is the population minimizer.
Hence, we have
\begin{align}
\label{equ:quadratic_Linear}
    0 \leq
    \frac{\ssmoothness}{2}\left[\left(1-\frac{1}{\condnum}\right)\norm{\Delta^{t-1}}^2_2 -
\left(1 - \frac{1}{\slack} \right)\norm{\Deltabar^t}^2_2 \right]+ \left( \alpha \norm{\Delta^{t-1}}_2 + \psi \right) \norm{\Deltabar^t}_2,
\end{align}
with probability at least $1-d^{-3}$.

Notice that
\cref{equ:quadratic_Linear} is a quadratic inequality for
$\norm{\Deltabar^t}_2$, and we can use the  root of \cref{equ:quadratic_Linear} to upper bound $\norm{\Deltabar^t}_2$:

\begin{align*}
    \norm{\Deltabar^t}_2 &\leq
    \frac{\left( \alpha \norm{\Delta^{t-1}}_2 + \psi \right) +
    \sqrt{\left( \alpha \norm{\Delta^{t-1}}_2 + \psi \right)^2
     + \left(\ssmoothness
    \left(1- \frac{1}{\slack}\right) \right) \cdot
    \ssmoothness \left(1  - \frac{1}{\condnum}\right)\norm{\Delta^{t-1}}_2^2
    }}{\ssmoothness
    \left(1- \frac{1}{\slack}\right)}  \\
   &\overset{(i)}{\leq}
    \frac{2 \left( \alpha \norm{\Delta^{t-1}}_2 + \psi \right)
    +     \norm{\Delta^{t-1}}_2
    \sqrt{\left(\ssmoothness
    \left(1- \frac{1}{\slack}\right) \right) \cdot
    \ssmoothness \left(1  - \frac{1}{\condnum}\right)
    }}{\ssmoothness
    \left(1- \frac{1}{\slack}\right) } \\
    &=
    \norm{\Delta^{t-1}}_2
    \sqrt{\frac{1  - \frac{1}{\condnum}
    }{ 1- \frac{1}{\slack} }}
    + \frac{2 \alpha \norm{\Delta^{t-1}}_2}{\ssmoothness
    \left(1- \frac{1}{\slack}\right) }
    +
    \frac{2 \psi}{\ssmoothness
    \left(1- \frac{1}{\slack}\right) }
\end{align*}
where (i) follows from the basic inequality $\sqrt{a+b}\leq \sqrt{a} + \sqrt{b}$
for non-negative $a, b$.

We choose $\slack = 2\condnum$,
and this leads to $ \sqrt{
\left(1- \frac{1}{\condnum}\right)/
\left( 1- \frac{1}{\slack} \right)} \leq 1 - \frac{1}{4\rho}$.
Under the condition
$\alpha \leq \frac{1}{32}\sconvexity$, we have $\frac{2 \alpha}{\ssmoothness \left(1- \frac{1}{\slack}\right) } \leq \frac{1}{8\condnum}$. Then,
\begin{align}
     \norm{\Deltabar^t}_2
     \leq
     \left( 1 - \frac{1}{8\rho} \right)\norm{\Delta^{t-1}}_2 +  \frac{4 \psi}{\ssmoothness
     }        \label{equ:recursion_linear}
\end{align}

Since $\param^{t+1} = \Pi_\constraint \left(\parambar^{t+1}\right)$
is projection onto a convex set, by the property of Euclidean projection \cite{bubeck2015convex}, we have
\begin{align}
    \norm{\Delta^t}_2 = \norm{\param^t - \param^*}_2 \leq \norm{\parambar^t - \param^*}_2 =
    \norm{\Deltabar^t}_2. \label{equ:euclidean}
\end{align}

Together with \cref{equ:euclidean},
\cref{equ:recursion_linear} establishes global linear convergence
of $\Delta^t$.

We
apply a union bound on $T$ iterates.
Since $  1-T d^{-3} \geq 1 - {d}^{-2}$ for sufficiently large $d$,
we have
\begin{align*}
\norm{\Delta^{T}}_2
\leq
  \left( 1 - \frac{1}{8\rho} \right)^{T}\norm{\param^*}_2 +  \frac{32 \psi}{\sconvexity}
\end{align*}
with probability at least $1 - {d}^{-2}$.
Hence, we can achieve the final error
\begin{align*}
     \norm*{\widehat{{\param}} - \param^*}_2 = O(\psi/\sconvexity),
\end{align*}
by setting
$ T = {O}\left(
\condnum \log\left(\frac{\sconvexity\norm{\param^*}_2}{\psi} \right)\right)$.

\fpro

  %

\subsection{Sparse linear regression}

\begin{proof}[Proof of \Cref{thm:main_linear} and \Cref{thm:main_linear_heavy}]

With \Cref{prop:heavy}, \Cref{prop:contamination} in hand, we prove
the arbitrary corruption case \Cref{thm:main_linear}, and the
proof of heavy tailed distribution \Cref{thm:main_linear_heavy}
is similar. 
We evaluate the RDC \Cref{def:RDC} in \Cref{alg:THT} for trimmed gradient estimator.
With probability at least $1-d^{-3}$, we have
\begin{align*}
    \abs{ \inner{\RobustG(\param) - {\UMean}(\param)}{\param^{t-1} - \param^*}}  
    &\overset{(i)}\leq 
    \norm{\RobustG(\param) - {\UMean}(\param)}_\infty \norm{\param^{t-1} - \param^*}_1\\
    &\overset{(ii)}\leq \sqrt{k'+k}
    \norm{\RobustG(\param) - {\UMean}(\param)}_\infty \norm{\param^{t-1} - \param^*}_2\\
    &\overset{(iii)}\leq \sqrt{k'+k}
\left( \sqrt{ \norm*{\param - \param^*}_2^2 + \sigma^2}
\left(
{ \eps \log (nd) } +
\sqrt{{\log d}/{n}} \right) \right) \norm{\param^{t-1} - \param^*}_2\\   
&\leq \sqrt{k'+k} \left(
{ \eps \log (nd) } +
\sqrt{{\log d}/{n}} \right)
\left( { \norm*{\param - \param^*}_2 + \sigma}
 \right) \norm{\param^{t-1} - \param^*}_2\\ 
    &\leq \left( \alpha \norm{\param - \param^*}_2 + \psi \right) \norm{\param^{t-1} - \param^*}_2,
\end{align*}
where (i) follows from Holder inequality, (ii) follows from the sparsity of ${\param^{t-1} - \param^*}$ in \Cref{alg:THT}, 
(iii) follows from plugging in \Cref{prop:contamination}, which yields $\alpha = \sqrt{k'+k} \left(
{ \eps \log (nd) } +
\sqrt{{\log d}/{n}} \right), \psi = \sigma \sqrt{k'+k} \left(
{ \eps \log (nd) } +
\sqrt{{\log d}/{n}} \right)$.

We
apply a union bound on $T$ iterates, and  $1-T d^{-3} \geq 1 - {d}^{-2}$ for sufficiently large $d$.
The condition
$\alpha \leq \frac{1}{32}\sconvexity$ in \Cref{thm:meta}
  can be achieved if
\begin{align*}
n = \Omega\left(\frac{ \condnum^2 k \log d}{\sconvexity^2} \right), \text{and } \eps =  O\left(\frac{\sconvexity}{ \condnum \sqrt{k} \log (nd)}\right).
\end{align*}
Since $\condnum = \ssmoothness/\sconvexity$, and
$\ssmoothness \geq 1$, these conditions can be expressed as
\begin{align*}
n = \Omega\left({ \condnum^4 k \log d} \right),  \text{and } \eps =  O\left(\frac{1}{ \condnum^2 \sqrt{k} \log (nd)}\right).
\end{align*}
The final error can be expressed as
\begin{align*}
\norm{\widehat{{\param}} - \param^*}_2
= O\left( { \condnum^2 \sigma }
\left(
\underbrace{\eps \sqrt{k} \log (nd) }_{{\substack{\text{robustness} \\ \text{error}}}} +
\underbrace{\sqrt{\frac{k \log d}{n}}}_{\substack{\text{statistical} \text{ error}}}
\right)\right).
\end{align*}
\end{proof}

\subsection{Sparse logistic regression}
%

\spro[Proof of \Cref{thm:main_logistic} and \Cref{thm:main_logistic_heavy}]
We prove \Cref{thm:main_logistic}, and the proof of \Cref{thm:main_logistic_heavy}
is similar.
With probability at least
$1 - {d}^{-3}$, we have
\begin{align*}
    \abs{ \inner{\RobustG(\param) - {\UMean}(\param)}{\param^{t-1} - \param^*}}
   &\overset{(i)}\leq \sqrt{k'+k}
\left( 
{ \eps \log (nd) } +
\sqrt{{\log d}/{n}} \right) \norm{\param^{t-1} - \param^*}_2\\
&\leq \left( \alpha \norm{\param - \param^*}_2 + \psi \right) \norm{\param^{t-1} - \param^*}_2,
\end{align*}
where (i) follows from the proof of \Cref{thm:main_linear} by using 
$
\norm*{\RobustG - \OGrad}_\infty = O\left( \sqrt{{\log d}/{n}} \right)$ in \Cref{prop:contamination}, and
$\alpha = 0, \psi =  \sqrt{k'+k} \left(
{ \eps \log (nd) } +
\sqrt{{\log d}/{n}} \right)$.


Similar to the proof in sparse linear regression,
this final error can be expressed as
\begin{align*}
\norm{\widehat{{\param}} - \param^*}_2
= O\left( { \condnum }^2
\left(
\underbrace{\eps \sqrt{k} \log (nd) }_{{\substack{\text{robustness} \\ \text{error}}}} +
\underbrace{\sqrt{\frac{k \log d}{n}}}_{\substack{\text{statistical} \text{ error}}}
\right)\right).
\end{align*}
\fpro

%% file: Appendix_Sparsity.tex
\section{Sparsity recovery
and sparse precision matrix estimation
}
\label{sec:appendix_sparsity}

\subsection{Sparsity recovery guarantee}

The same as the main text, we use $\supp(\vect, k)$ to denote top $k$ indexes of $\vect$ with the largest magnitude.      
Let $\vect_\mathrm{min}$ denote the smallest absolute value of nonzero elements of $\vect$.
%
%
\spro[Proof of \Cref{thm:sparsity}]
The sparsity recovery guarantee is similar to \cite{GradientHT2018}.
Since $\widehat{\param}$ is $k'$ sparse ($k' \geq k$) by the definition of hard thresholding operator, we use $\widehat{\param}_k$ to denote $\Hard{k} (\widehat{\param})$.
We use the technique proof by contradiction.
If $\supp(\widehat{\param}, k) \neq \supp({\param^*})$, we at least have $\ell_2$ error as $\param_\mathrm{min}^*$. Hence,
$\param_\mathrm{min}^* \leq \norm{\widehat{\param}_k - \param^*}_2 \overset{(i)}{\leq} 2 \norm{\widehat{\param} - \param^*}_2  \overset{(ii)}{=}
O\left( { \condnum }^2 \sigma
\left(
 \eps \sqrt{k} \log (nd)  +
\sqrt{\frac{k \log d}{n}} \right)\right)$,
where (i) follows from the triangle inequality
and definition of hard thresholding
$\norm*{\widehat{\param}_k - \param^*}_2 \leq \norm*{\widehat{\param}_k - \widehat{\param}}_2  + \norm*{ \widehat{\param} - {\param}^*}_2 \leq 2 \norm*{\widehat{\param} - \param^*}_2$, and (ii) follows from the statistical guarantee in \Cref{thm:main_linear}.

This contradicts with the $\param_\mathrm{min}$-condition in \Cref{thm:sparsity}, and hence we have the result in  \Cref{thm:sparsity}.
\fpro

\subsection{Model selection for Gaussian graphical models}

We then start to consider the sparsity recovery results for sparse precision matrix estimation -- this is the part of \Cref{cor:sparsity}. We first use following notations for a Gaussian graphical model.

We use $\bm{x}_i$ to denote the $i$-th samples of Gaussian graphical model, and $X_j$ to denote the
$j$-th random variable.
Let $(j)$ be the
index set $\{1, \cdots , j- 1, j + 1, \cdots , d\}.$
We use
$\NeiCov = \bm{\Sigma}_{(j), (j)} \in \Real^{(d-1)\times(d-1)}$ to denote the sub-matrix of covariance matrix
$\bm{\Sigma}$ with both $j$-th row and $j$-th  column removed,
and use  $\NeiVec \in \Real^{d-1}$  to denote $\bm{\Sigma}$'s $j$-th column with the diagonal entry removed.
Also, we use $\bm{\theta}_{(j)} \in \Real^{d-1}$ to denote $\precision$'s $j$-th column with the diagonal entry removed.
and $\bm{\Theta}_{j,j} \in \Real$ to denote the $j$-th diagonal element of $\bm{\Theta}$.

By basic probability computation,
for each $j = 1, \cdots, d$, the variable $X_j$ conditioning  $\NeiRV$  follows from a Gaussian distribution
$\mathcal N (\NeiRV^\top \NeiCov^{-1} \NeiVec, 1 -  \NeiVec^\top  \NeiCov^{-1} \NeiVec)$.
Then we have the linear regression formulation
$    X_j = \NeiRV^\top \param_j + \xi_j$,
where $\param_j = \NeiCov^{-1} \NeiVec$
and $\xi_j \sim \mathcal{N} (0,  1 -  \NeiVec^\top  \NeiCov^{-1} \NeiVec)$.
Notice the definition of precision matrix
$\precision$, we have
$\param_j  = -\bm{\theta}_{(j)}
/\precision_{j,j}$, and $\precision_{j,j} = 1/\Var(\xi_j )$.
Thus for the $j$-th variable, $\bm{\theta}_{(j)}$ and $\param_j$ have
the same sparsity pattern.
Hence, the sparsity pattern of $\bm{\theta}_{(j)}$ can be estimated through $\widehat{\param}_j$
via solving the optimization \cref{equ:NS} (Neighborhood Selection in \cite{meinshausen2006high}).

\begin{algorithm}[t]
\begin{algorithmic}[1]
\STATE \textbf{Input:} Data samples $\{\bm{x}_i\}_{i=1}^m$.
\STATE \textbf{Output:} The sparsity pattern estimation of  $\precision$.
\STATE \textbf{Parameters:} Hard thresholding parameter $k'$.\\
{\kern2pt \hrule \kern2pt}
\FOR{each variable $j$,}
\STATE Use $X_j$ as response variable, and   $\NeiRV$ as covariates.
\STATE Run \Cref{alg:THT} with input $\{x_{ij}, \bm{x}_{i(j)}\}_{i=1}^m$. We set the parameter as $k'$, and the loss function as least square loss, and use trimmed gradient estimator.
\STATE The output of \Cref{alg:THT} is denoted as $\widehat{\param}_j\in \Real^{d-1}$.
\ENDFOR
\STATE Aggregate the neighborhood support set of $\{\widehat{\param}_j\}_{j=1}^d$ via intersection or
union.
\end{algorithmic}
\caption{Neighborhood Selection via Robust Hard Thresholding (Robust NS)}
\label{alg:graphical_IHT}
\end{algorithm}

In \Cref{alg:graphical_IHT}, we robustify Neighborhood Selection by
using Robust Hard Thresholding (with $\ell_2$ loss and trimmed gradient estimator) to robustify \cref{equ:NS}.
In line 6, we use Robust Hard Thresholding to regress
each variable against its neighbors.
 In line 9, the sparsity pattern of $\precision$ can be estimated by aggregating the neighborhood support set of $\{\widehat{\param}_j\}_{j=1}^d$ via intersection or
union.
Similar to  \Cref{thm:sparsity},
a $\bm{\theta}_\mathrm{min}$-condition guarantees consistent edge selection.
%

\spro[Proof of \Cref{cor:sparsity}]
\Cref{alg:graphical_IHT} iteratively uses \Cref{alg:THT} as a Neighborhood Selection approach for each variable. Hence, we can apply \Cref{thm:sparsity} for each variable, and the sparsity patterns are the same according to
$\bm{\theta}_{(j)} = -\param_j / \Var(\xi_j)$.
The stochastic noise term $\sigma$ in sparse linear regression can be expressed as $1/\sqrt{\precision_{j,j}}$.
Hence, under the same condition as \Cref{thm:main_linear}, for each $j\in[d]$,
we require a
$\bm{\theta}_\mathrm{min}$-condition for $\bm{\theta}_{(j)}$,
$    \bm{\theta}_{(j),\mathrm{min}} =
    \Omega\left( { {\precision_{j,j}^{1/2}} \condnum^2 }\left({ \eps \sqrt{k} \log (nd) } +
\sqrt{\frac{k \log d}{n}} \right)\right)$.

Using a union bound,
we conclude that \Cref{alg:graphical_IHT} is consistent
in edge selection, with probability at least $1 - d^{-1}$.
\fpro

%% file: Additional.tex
\section{Full experiments details}

\label{sec:additional}

\input{Experiments_appendix.tex}

%% file: Experiments_appendix.tex
We study empirical performance
of Robust Hard Thresholding (\Cref{alg:THT} and \Cref{alg:graphical_IHT}). 
And we present the complete details of experimental setup in \Cref{sec:experiments}.

\subsection{Synthetic data -- sparse linear models}

We first consider the performance of \Cref{alg:THT} under (generalized) linear models
with $\eps$-corrupted samples.

\textbf{Sparse linear regression. }
In the first experiemtn, we consider 
an exact sparse linear regression model (\Cref{model:linear}).
In this model, the stochastic noise $\xi \sim \mathcal{N}(0, \sigma^2)$, and we vary the noise level $\sigma^2$ in different simulations. 
We first generate
authentic explanatory variables with  
parameters $k = 5, d = 1000, n = 300$, from a Gaussian distribution
$\mathcal{N} (\bm{0}_d, \bm{\Sigma})$, where the covariance matrix  $\bm{\Sigma}$ is a Toeplitz matrix with an
exponential decay $\bm{\Sigma}_{ij} = \exp^{-|i-j|}$.
This design matrix is known to enjoy the RSC-condition \cite{raskutti2010restricted}, which meets the requirement
of \Cref{thm:main_linear}.
The entries of the $k$-sparse true parameter $\param^*$ are set to either $+1$ or $-1$.
Fixing the contamination level at $\eps = 0.1$,
we  set the covariates of the outliers as $A$, where $A$ is a random $\pm 1$ matrix of dimension $\frac{\eps}{1 - \eps} \times d$, and the responses of outliers to $-A\param^*$.

\begin{figure}[t]
\centering
\subfloat[][$\log(\norm{{\param^t} -\param^*}_2)$  vs. iterations  under \Cref{model:linear}.]{
\includegraphics[width=.4\columnwidth]{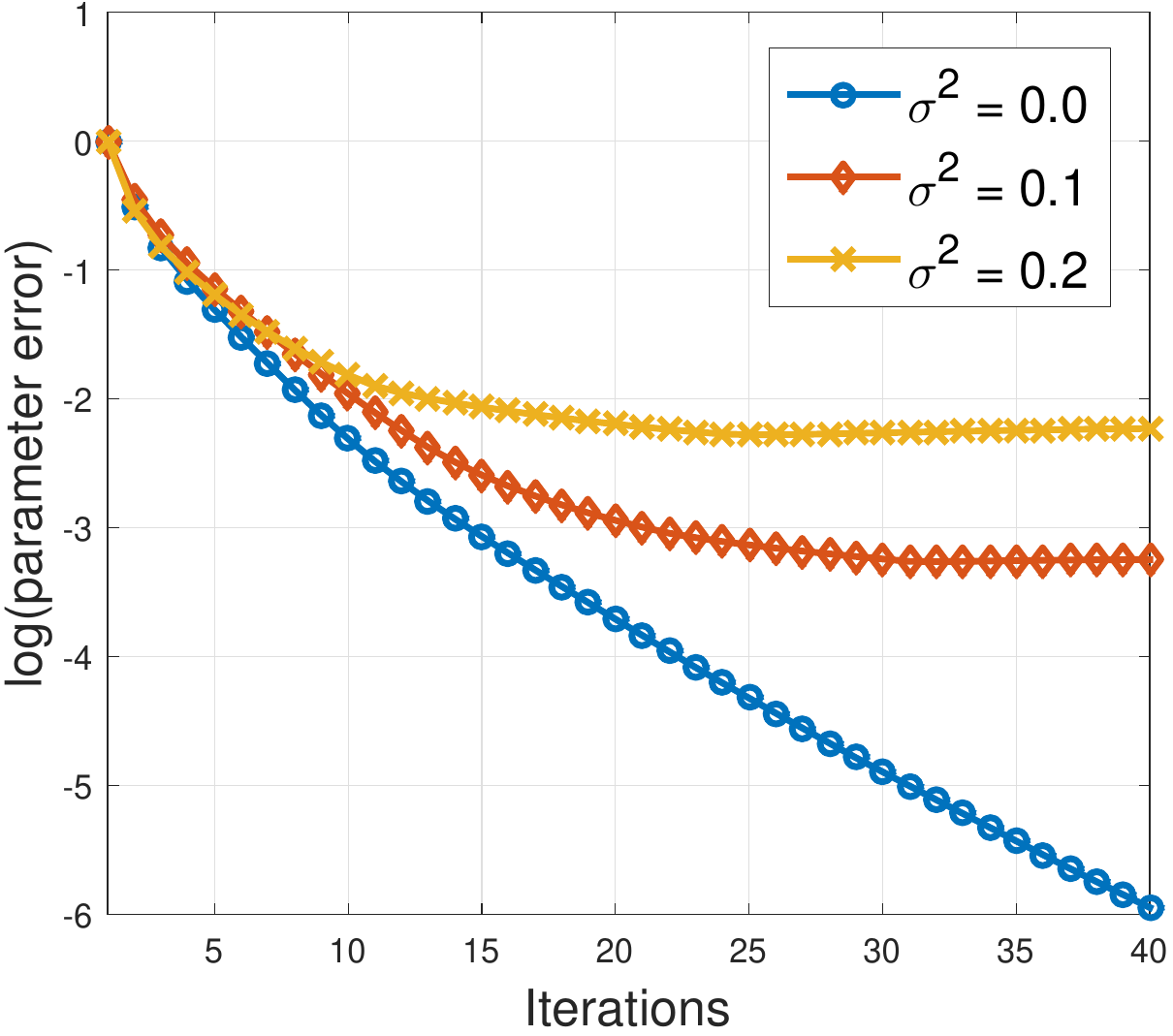}}$\quad$
\subfloat[][Function value  vs. iterations  under model misspecification.]{
\includegraphics[width=.4\columnwidth]{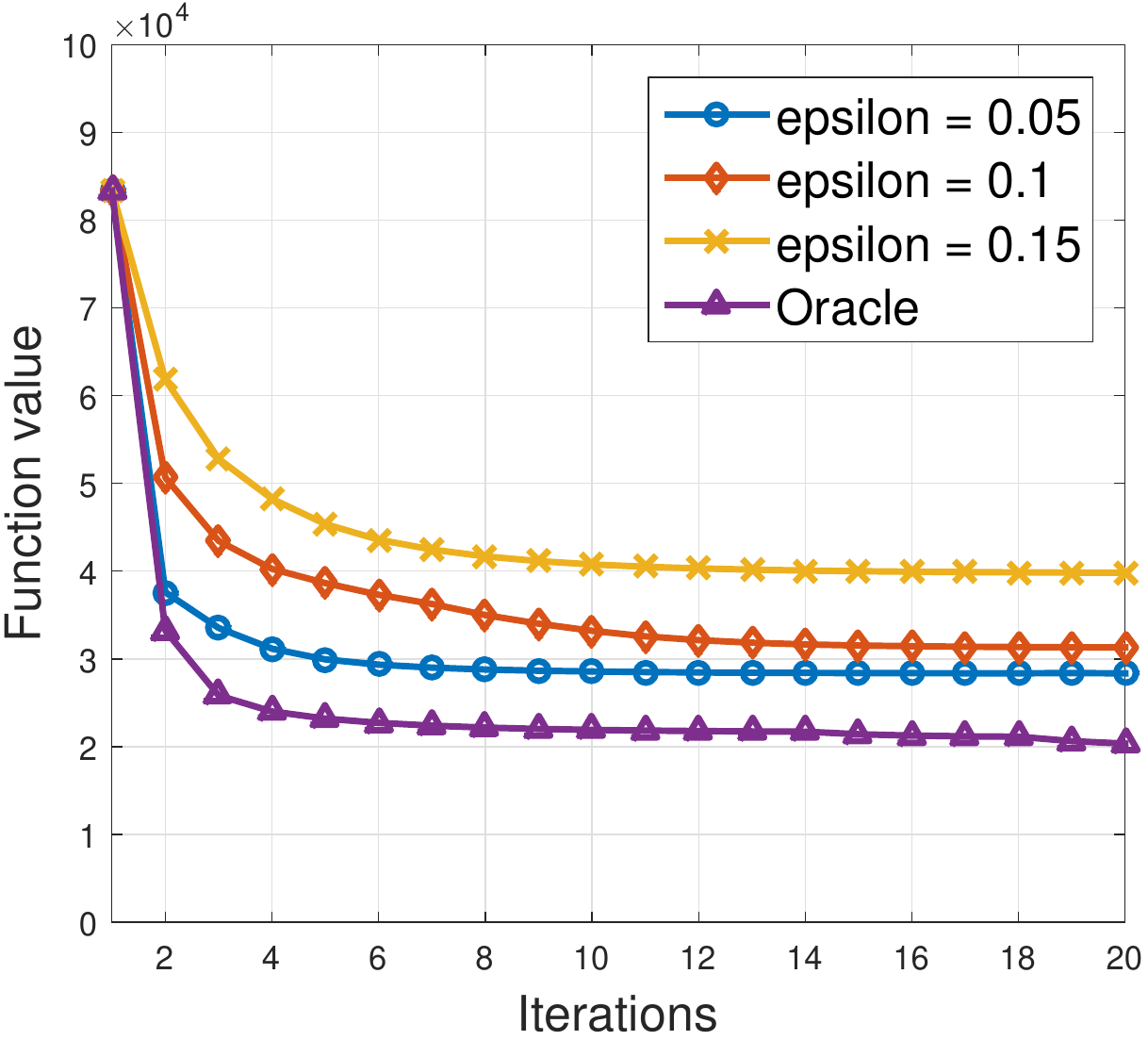}}
\caption{\footnotesize{
In the left plot, we use
\Cref{alg:THT} under \Cref{model:linear} with different noise level $\sigma^2$.
In the right plot, we use
\Cref{alg:THT} under model misspecification with $\eps$. 
The function value is defined as
$F(\param) = \sum_{i \in \Good}(y_i - \bm{x}_i^\top \param)^2$.
}}
\label{fig:app:Linear}
\end{figure}

To show the performance of \Cref{alg:THT} under different noise levels determined by $\sigma^2$,
we track the parameter error $\norm{{\param^t} -\param^*}_2$ in each iteration.
In the left plot of \Cref{fig:app:Linear},
 \Cref{alg:THT} shows linear convergence, and the error curves
flatten out at the level of the final error, which is consistent with our theory.
Furthermore,  \Cref{alg:THT} can  achieve  machine precision when $\sigma^2 = 0$, which means exact recovery of $\param^*$.

\textbf{Misspecified model. }
For the second experiment, we use a sparse linear regression with model misspecification -- the underlying authentic samples do not follow   a linear model.
We use the same Toeplitz covariates and true parameter $\param^*$, but
and corresponding $y_i$'s are calculated as $y_i = \sum_{j=1}^{d}\bm{x}_{ij}^{3}\param_j^*$.
Although this is a non-linear function,  sparse linear regression on these authentic samples
can still recover the support, as the cubic function is monotone and $\param^*$ is sparse.
We generate  outliers using the same distribution as the first experiment, but with a different fraction of corruptions $\eps$.

For simplicity, we track the function  evaluated on all authentic samples 
$F(\param) = \sum_{i \in \OGood}(y_i - \bm{x}_i^\top \param)^2$.
In the right plot of \Cref{fig:app:Linear}, we show the performance of \Cref{alg:THT} under different
$\eps$, and the oracle curve means using IHT \emph{only on} authentic samples.
The right plot has similar convergence under different values of corrupted fraction $\eps$, and shows
the robustness of
\Cref{alg:THT} without assuming an underlying linear model.

\subsection{Robust $M$-estimators via Robust Hard Thresholding}
Classical robust $M$-estimators \cite{loh2017statistical} (such as empirical risk minimization using Huber loss) are widely used in robust statistics in the case where the error distribution is heavy tailed or when there are arbitrary outliers {\em only in the response variables}.
In the high dimensional setting, 
given $\eps$-corrupted samples \Cref{def:contamination_model},
we can use 
\begin{align*}
   \min_{\param \in \Omega } \Expe_{i \in_u \Input} \loss_i(\param; \bm{z}_i), \quad \text{s.t.}
   \norm{\param}_0 \leq k,
\end{align*}
where $\loss_i(\param; \bm{z}_i)$ can be chosen as Huber loss   with parameter $\delta$:
\begin{align*}
H_{\delta }(\param; \bm{z}_i)={\begin{cases}{\frac  {1}{2}}(y_i- \bm{x}_i^\top \param)^{2}&{\textrm  {for }} \abs{y_i - \bm{x}_i^\top \param} \leq \delta ,\\\delta \,\abs{y_i- \bm{x}_i^\top \param}-{\frac  {1}{2}}\delta ^{2}&{\textrm  {otherwise.}}\end{cases}}
\end{align*}

\cite{loh2017statistical} studied
robust $M$-estimators in high dimensions, and proposed 
a composite optimization using $\norm{\param}_1$ instead of $\norm{\param}_0$.
They established local convergence guarantee for this composite optimization procedure,  using a local RSC condition in a neighborhood around $\param^*$.
Yet their results do not 
trivially extend to settings with arbitrarily corrupted covariates.

In our experiments, we use Huber loss   in Robust Hard Thresholding to deal with heavy-tailed error distribution.
In addition to heavy-tailed noise,   $\eps$-fraction of  $\{y_i, \bm{x}_i\}_{i=1}^n$ are
still arbitrarily corrupted.

For the experiments, we use the same Toeplitz covariates and true parameter $\param^*$ as in previous experiments on sparse linear models with fixed dimension parameters $k = 5, d = 1000, n = 300$. The error distribution is a Cauchy distribution, which is a special case  model misspecification, as it doesn't meet the sub-Gaussian requirement in \Cref{model:linear}.
For different contamination levels,
we   set the covariates of the outliers as $A$, where $A$ is a random $\pm 1$ matrix of dimension $\frac{\eps}{1 - \eps} \times d$, and the responses of outliers to $-A\param^*$.

\begin{figure}[t]
\centering
\includegraphics[width=.4\columnwidth]{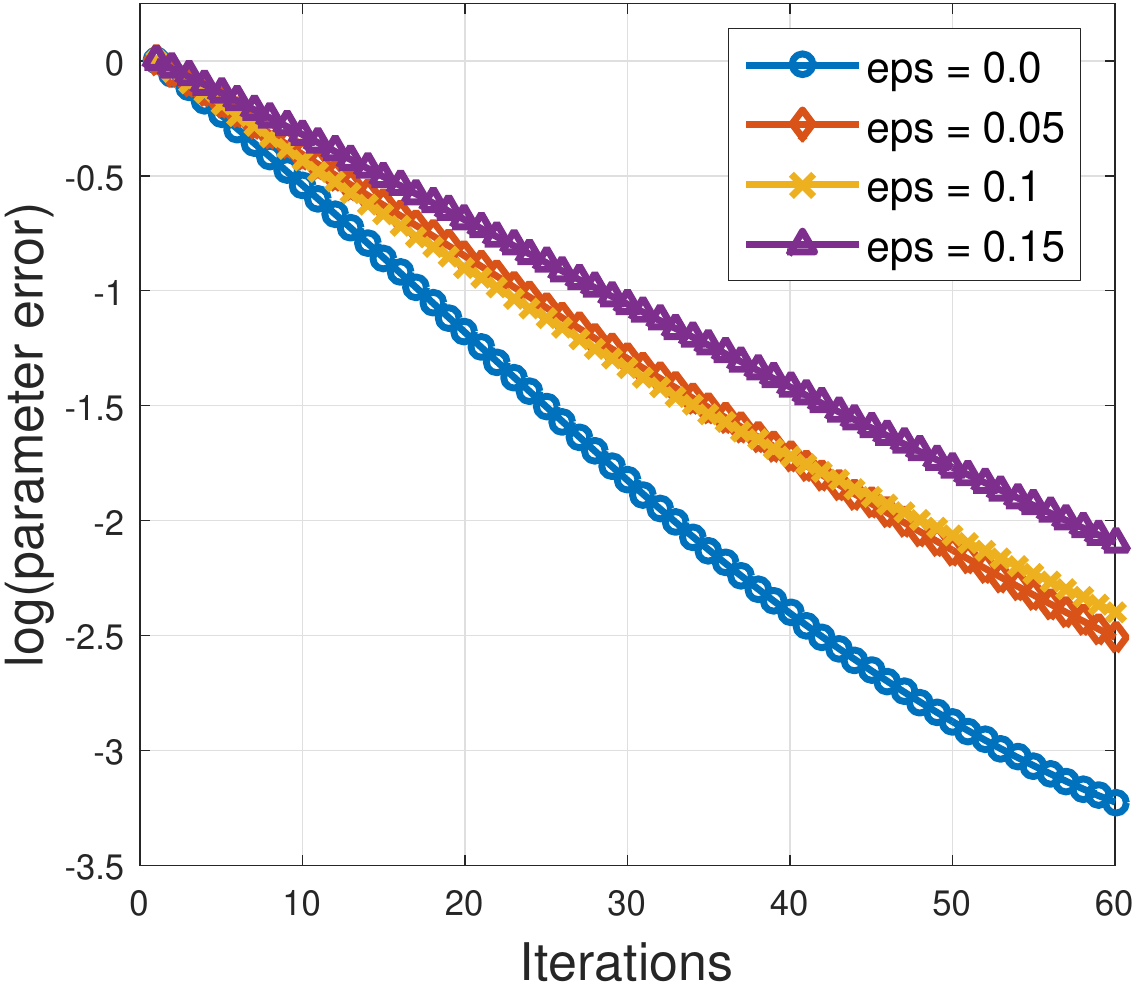}
\caption{
{ \footnotesize
$\log\| \param^t -\param^* \|_2$
 vs. iterations for Robust Hard Thresholding using Huber loss in the  sparse linear model where the error distribution is a heavy-tailed Cauchy distribution.}
}
\label{fig:cauchy}
\end{figure}

Empirically, we  observe linear convergence, and this is shown in \Cref{fig:cauchy}. 
This linear convergence results validates the local RSC condition proposed in \cite{loh2017statistical}, and we can still achieve this even with  $\eps$-fraction of corrupted covariates.

\subsection{Sparse logistic regression}

For binary classification problem,
we generate samples from a sparse LDA problem,
where the  distributions of the explanatory variables conditioned on the response variables follow
 multivariate Gaussian distributions
with the same covariance matrix but different means.

We generate authentic samples $\bm{x}_i$ from a Gaussian distribution $\mathcal{N} (\bm{\mu}_{+}, \bm{I}_d)$ if $y_i = +1$, and another distribution $\mathcal{N} (\bm{\mu}_{-}, \bm{I}_d)$
if $y_i = -1$.
The   parameters are fixed $k = 5, d = 1000, n = 300$.
We set $\bm{\mu}_{+} = \bm{1}_d + \bm{v}$, where
$\bm{v}$ is $k$-sparse
and its entries are set to be either $+1/\sqrt{k}$ or $-1/\sqrt{k}$.
And we set $\bm{\mu}_{-} = \bm{1}_d - \bm{v}$. 
The Bayes classifier is
$\param^* = 2 \bm{v}$.
This is a special case of \Cref{model:logistic}, and
it is known that sparse logistic regression attains fast classification error rates \cite{li2015fast}.
We then set the covariates of the outliers as $A$, where $A$ is a matrix of dimension $\frac{\eps}{1 - \eps} \times d$, where the entries are random $\pm 3$. The responses of outliers follow the distribution
$\Pr(y_i|\bm{x}_i) = {1}/({1 + \exp(y_i \bm{x}_i^\top \param^*)})$, which is exactly the opposite of  \Cref{model:logistic}.

\begin{figure}[t]
\centering
\subfloat[][$\log(\norm{{\param^t} -\param^*}_2)$  vs. iterations under \Cref{model:logistic}.]{
\includegraphics[width=.4\columnwidth]{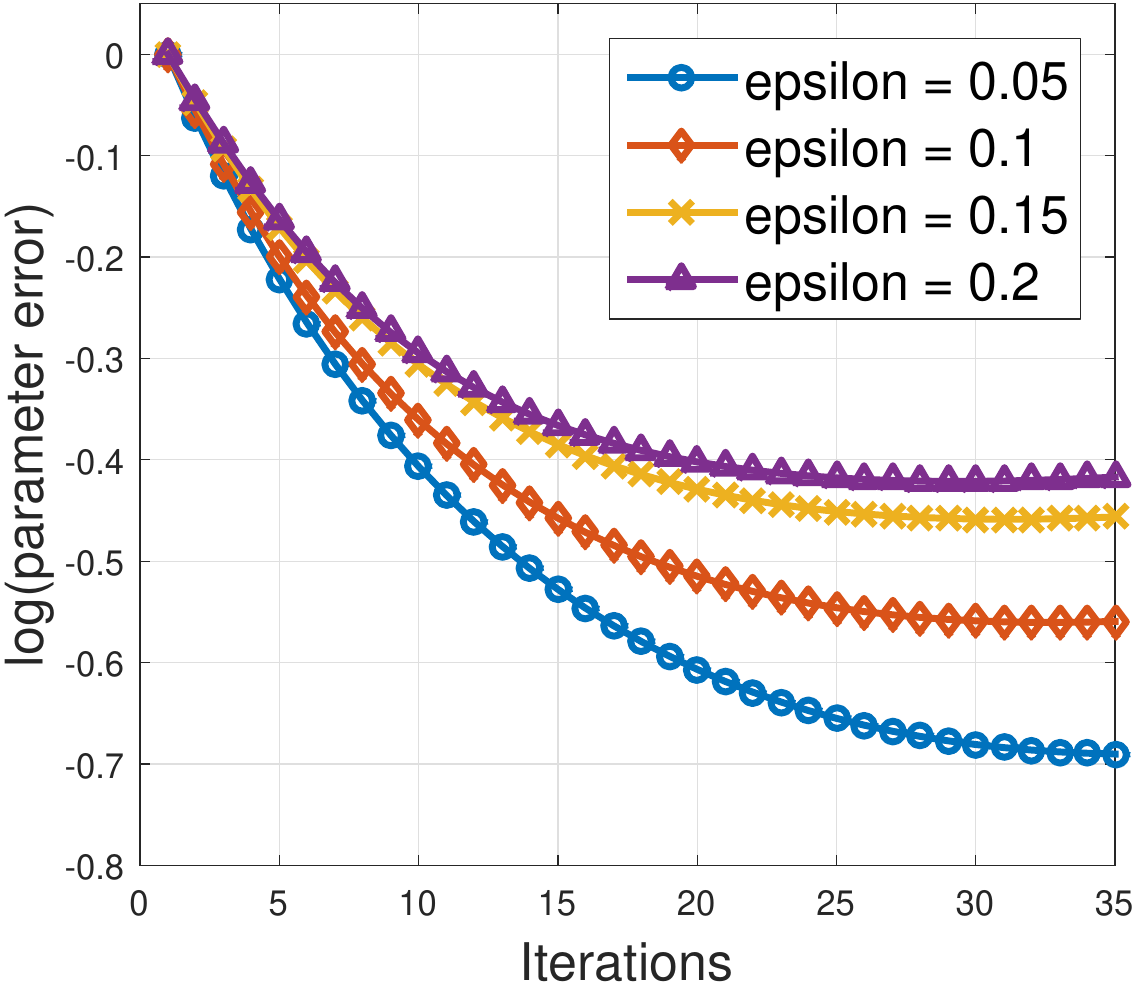}}$\quad$
\subfloat[][Classification error vs. dimensions under different $\eps$.]{
\includegraphics[width=.4\columnwidth]{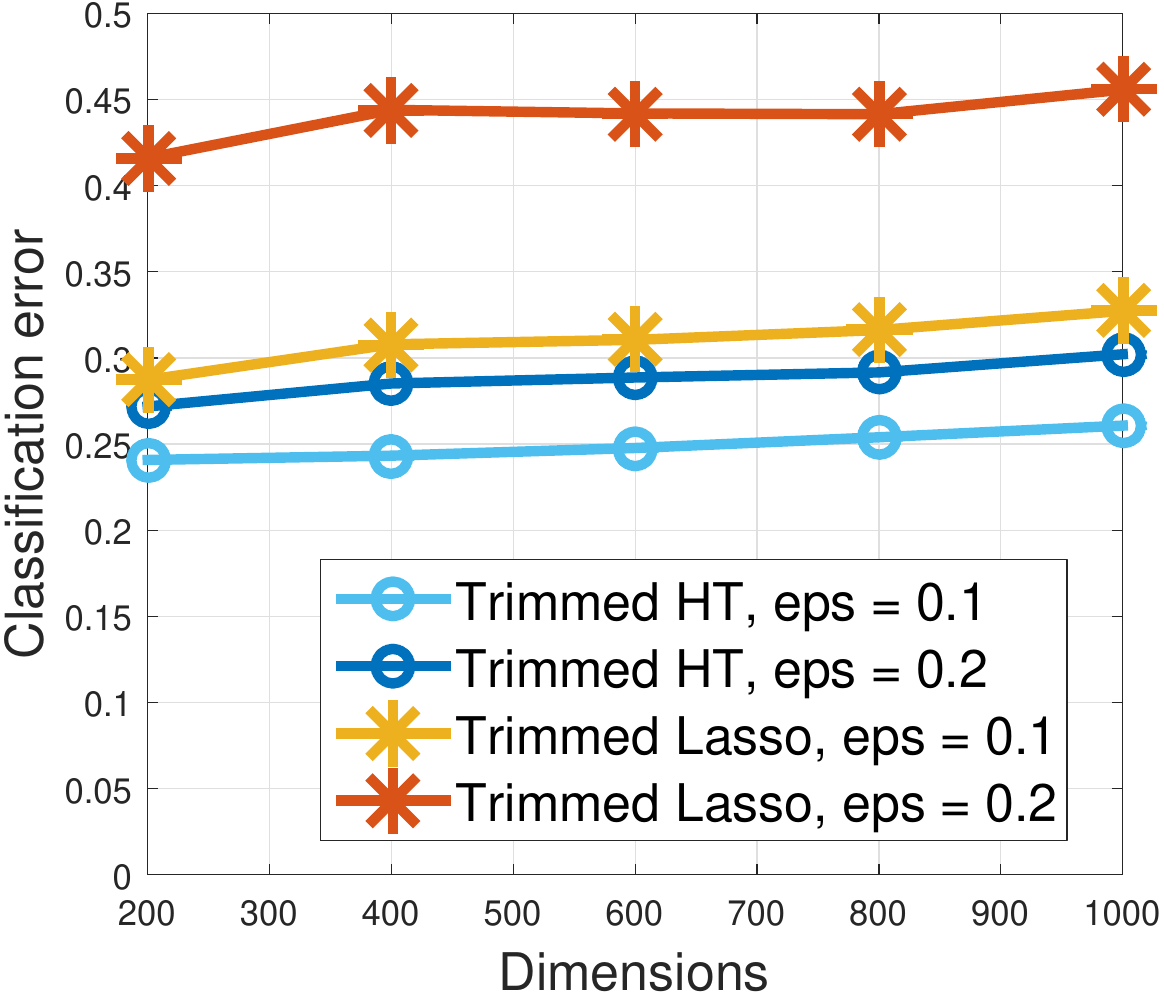}}
\caption{\footnotesize{The left plot demonstrates linear convergence of \Cref{alg:THT} for sparse logistic regression.
The right plot compares  our Robust Hard Thresholding (denoted as Trimmed HT) to Trimmed Lasso  \cite{Trim2018general}
in binary classification problem.}}
\label{fig:app:Logistic}
\end{figure}

We run \Cref{alg:THT} with logistic loss under different levels of outlier fraction $\eps$.
In the left plot of \Cref{fig:app:Logistic}, we observe similar linear convergence as
sparse linear regression This is consistent with \Cref{thm:main_logistic} for sparse logistic regression, and
it is clear that we cannot exactly recover $\param^*$ unless the number of samples $n$ is infinite.

We then compare \Cref{alg:THT} with the Trimmed Lasso estimator for sparse logistic regression   \cite{Trim2018general}.
Although they also use a trimming technique, their algorithm is totally different from \Cref{alg:THT}, as we use
coordinate-wise trimmed mean estimator for gradients in hard thresholding, but they trim samples in each iteration according to the each sample's loss. Under the same sparse LDA model, we set $k = \sqrt{d}, n = 15k$. In simulation, we increase $d$, and plot
classification error (averaged over 50 trials on authentic test set) for different $\eps=0.1, 0.2$.
The right plot of \Cref{fig:app:Logistic} shows that
Robust Hard Thresholding is better than Trimmed Lasso.

\subsection{Synthetic data -- Gaussian graphical model}
\label{sec:Huge}
\begin{figure}[t]
\centering
\subfloat[][$d=100$, low SNR]{
\includegraphics[width=.4\columnwidth]{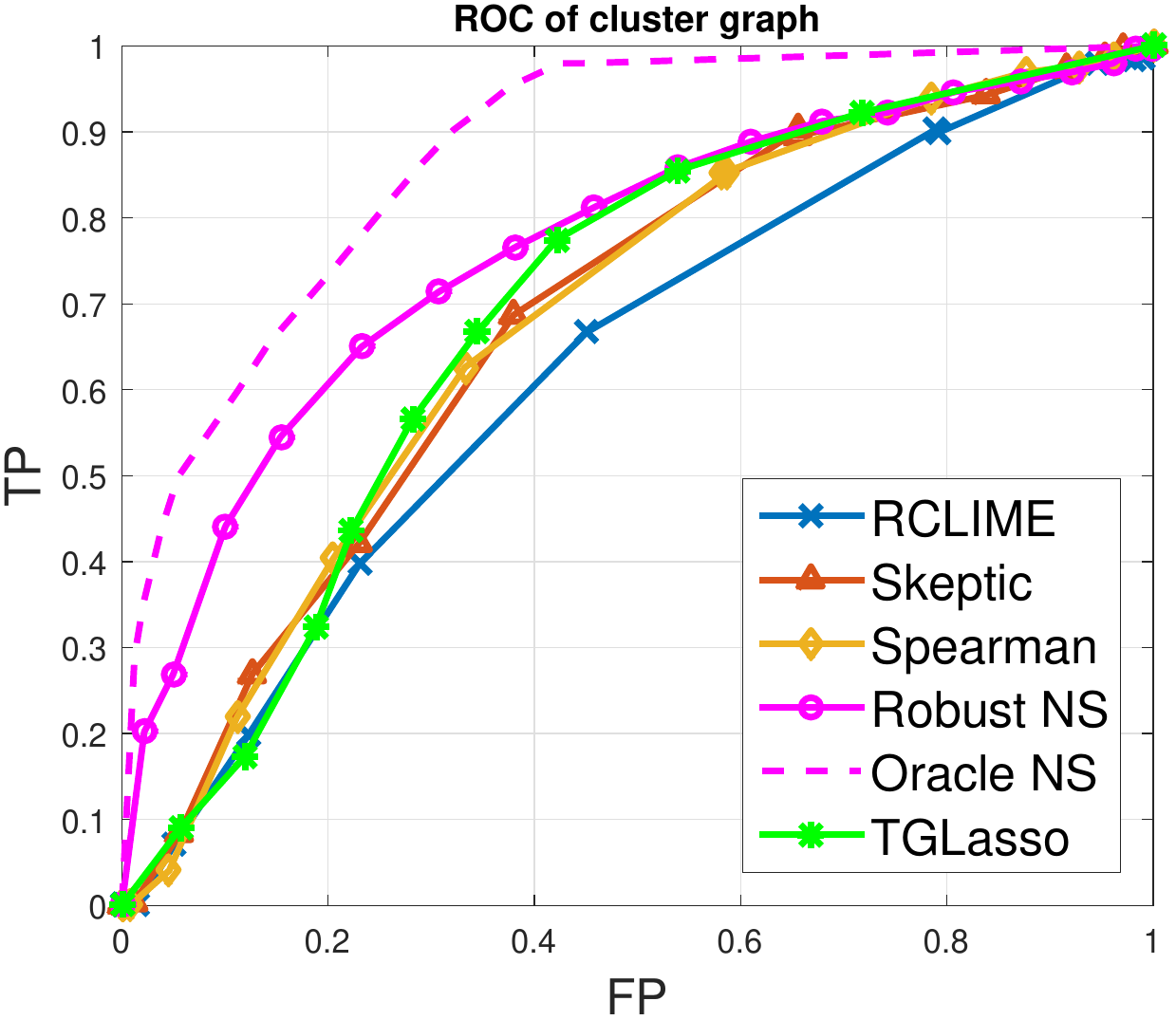}}
\subfloat[][$d=100$, high SNR]{
\includegraphics[width=.4\columnwidth]{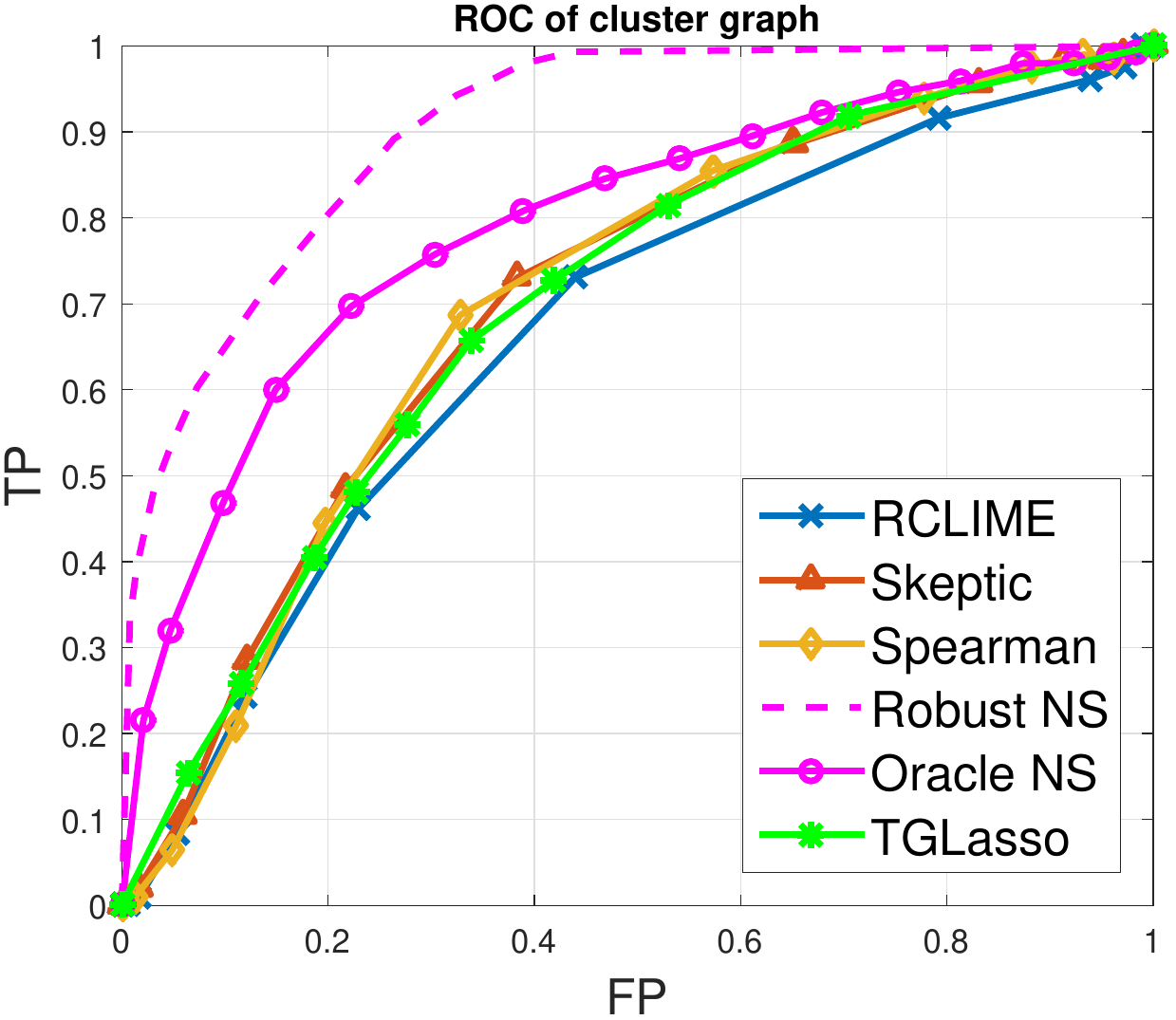}}

\subfloat[][$d=200$, low SNR]{
\includegraphics[width=.4\columnwidth]{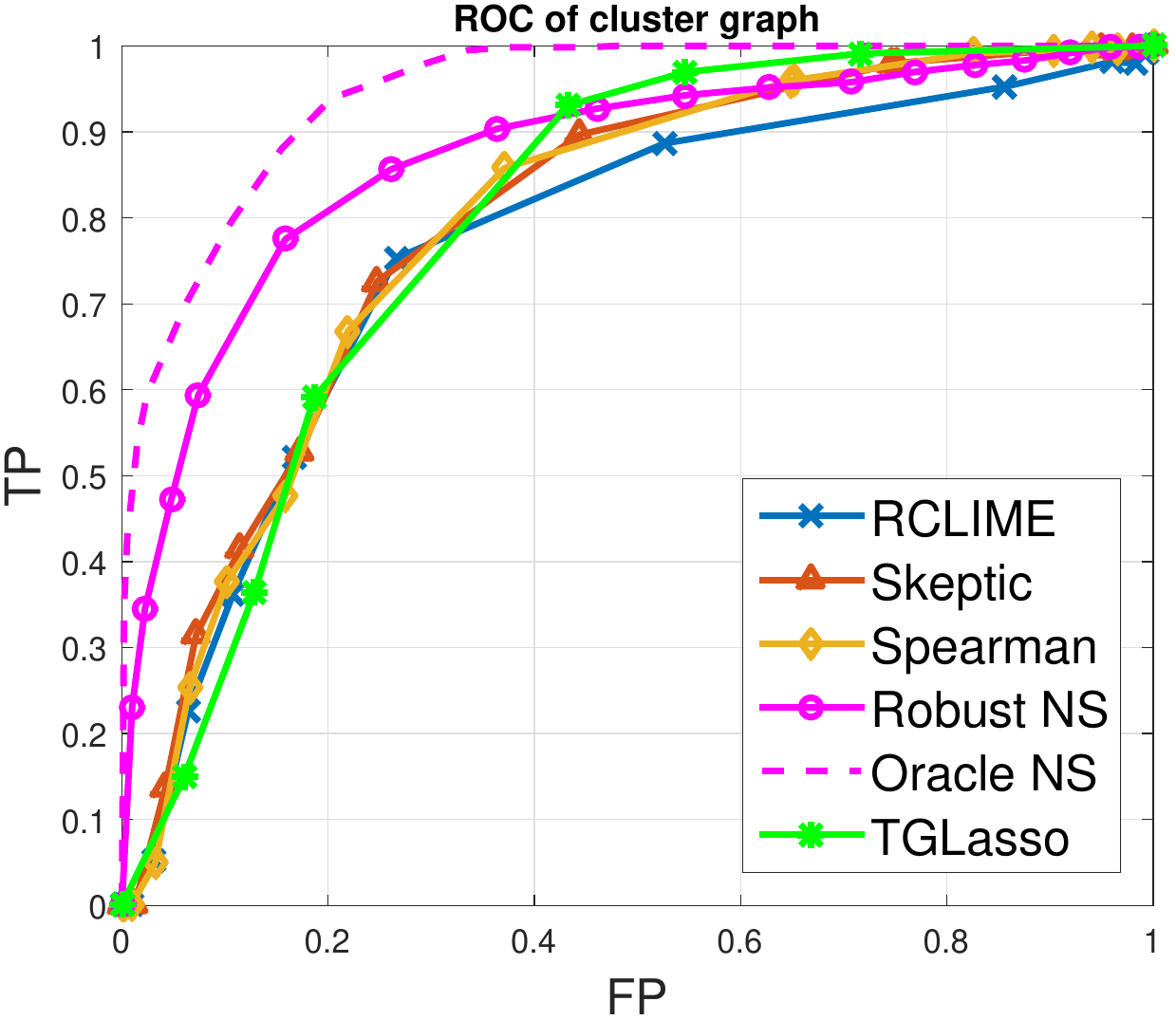}}
\subfloat[][$d=200$, high SNR]{
\includegraphics[width=.4\columnwidth]{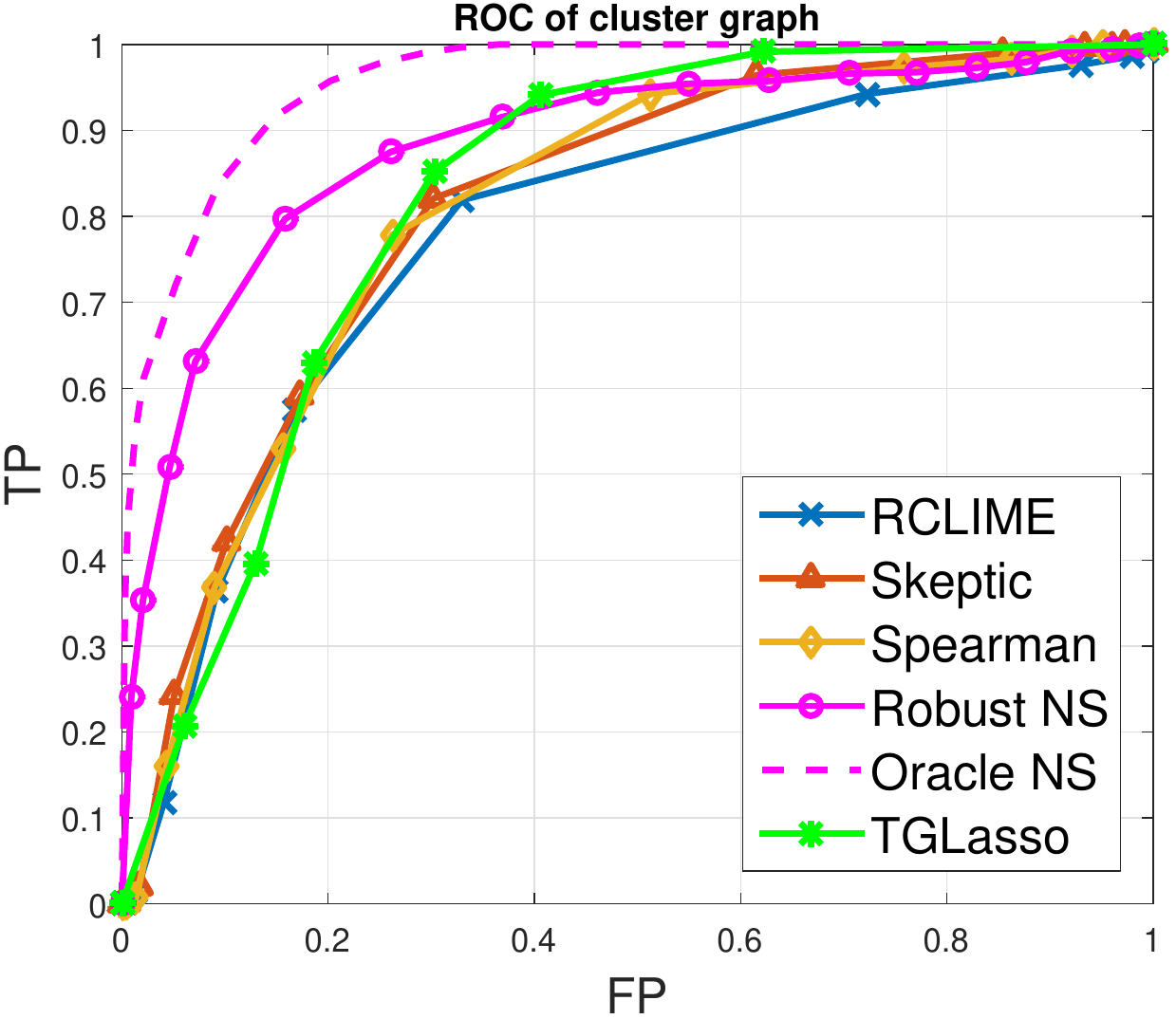}}
\caption{\footnotesize{ROC curves of different methods on cluster graphs with arbitrary corruptions. The curve Robust NS denotes  \Cref{alg:graphical_IHT}, and Oracle NS denotes the neighborhood selection Lasso only on authentic data.  
}}
\label{fig:app:Robust_Graphical}
\end{figure}

We generate Gaussian graphical model samples by \texttt{huge}  
\cite{HugePackage}.
We choose the ``cluster'' sparsity pattern, where the clustering parameters are default values in the package where the number of clusters in the graph is $d/20$,
 the probability that a pair of
nodes within a  cluster are connected
is 0.3, and there are no edges between nodes within different clusters. 
The off-diagonal elements of the precision matrix is denoted as $v$, which is an experiment parameter for SNR.

We then add an additional
$\frac{\eps}{1 - \eps}$
fraction of samples sampled from another   distribution.
Following the experimental design in \cite{Trim2015glasso, gu2017robust},
each outlier is generated by a mixture of $d$-dimensional Gaussian distributions $\frac{1}{2}\mathcal{N} (\bm{\mu}^o, \bm{\Sigma}^o) + \frac{1}{2}\mathcal{N} (-\bm{\mu}^o, \bm{\Sigma}^o)$, where
$\bm{\mu}^o = (1.5, 1.5, \cdots, 1.5)^\top, \text{ and } \bm{\Sigma}^o = \bm{I}_d.$
We compare \Cref{alg:graphical_IHT} with other existing methods:
Trimmed GLasso \cite{Trim2018general},
RCLIME \cite{gu2017robust},
Skeptic \cite{liu2012Skeptic},
and Spearman \cite{loh2018Spearman}.
The latter two are based on robustifying  the covariance matrix, and then using standard graphical model selection algorithms such as GLasso or CLIME. To directly compare these methods, we use CLIME for both of them.

To evaluate   model selection performance, we use receiver operating characteristic (ROC) curves to compare  our method to others   over the full regularization paths.
We generate   regularization paths for other robust algorithms by tuning  the $\lambda$ in CLIME and  GLasso.
For \Cref{alg:graphical_IHT}, we explicitly tune different sparsity level
$k'$ to generate the regualization path.

We set  $\eps = 0.1$, and vary  $(n, d)$, and the SNR parameter $v$ for
off-diagonal elements.
We use different  $(n, d) = (100, 100), (200, 200)$.
For different off-diagonal values, we set 
$v = 0.3$ (Low SNR), and 
$v = 0.6$ (High SNR).
We show ROC curves to demonstrate   model selection performance in
\Cref{fig:app:Robust_Graphical}. 
For the entire  regularization path, our algorithm (denoted as Robust NS) has a better ROC compared to other algorithms.

In particular, Robust NS outperforms other methods with higher true positive rate
when the false positive rate is small.         
This is the case where we use smaller
hard thresholding sparsity
in \Cref{alg:graphical_IHT},
and larger regularization parameter for $\norm{\Theta}_1$ other methods based on GLasso and CLIME.
This validates our theory in \Cref{cor:sparsity}, which guarantees sparsity recovery   when hard thresholding hyper-parameter $k'$  is suitably chosen to match $\param^*$'s sparsity $k$.

\subsection{Real data experiments}

\begin{figure}[t]
\centering
\subfloat[][An example from the sector Information Technology.]{
\includegraphics[width=.4\columnwidth]{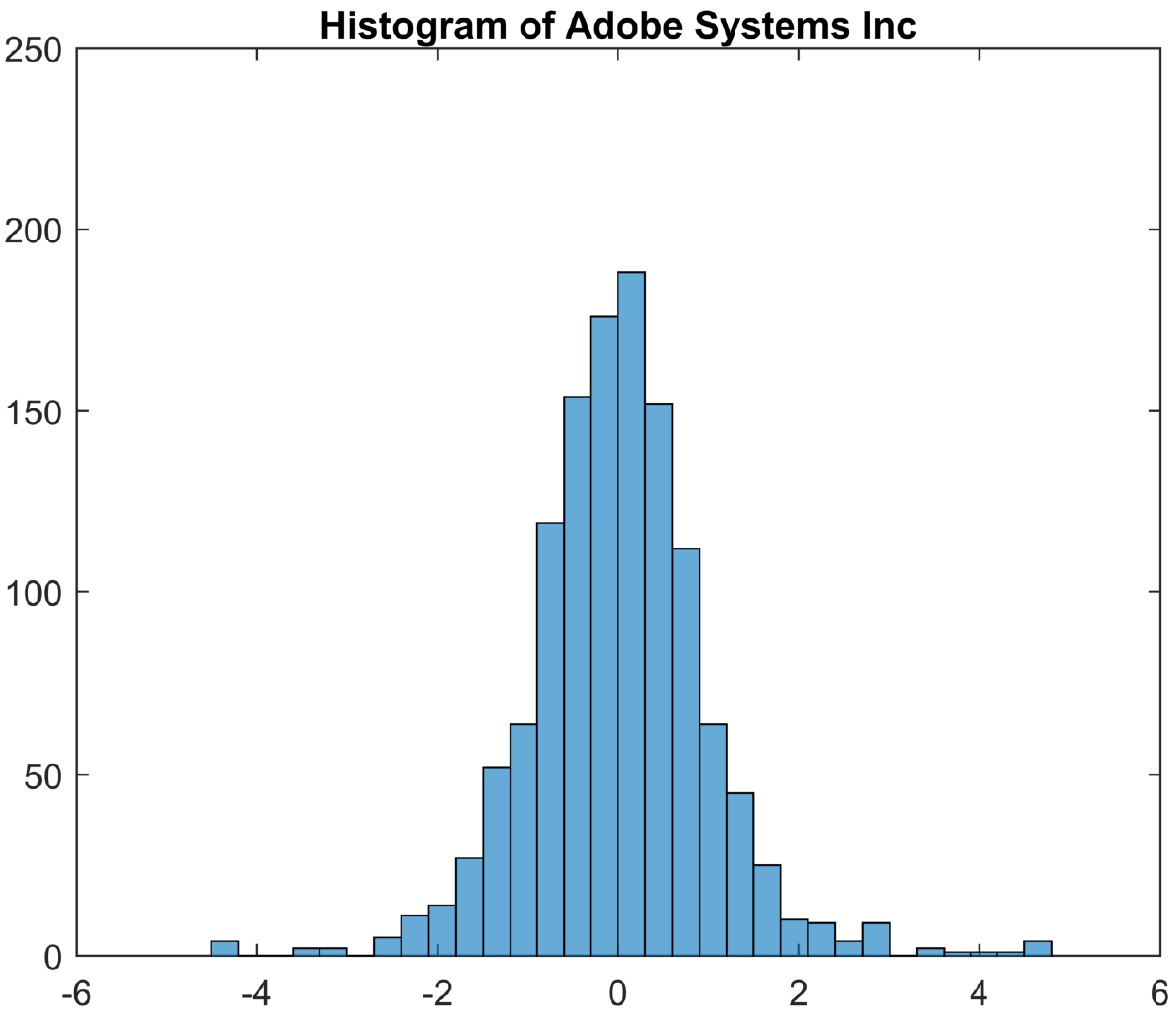}}$\quad$
\subfloat[][An example from the sector Information Technology.]{
\includegraphics[width=.4\columnwidth]{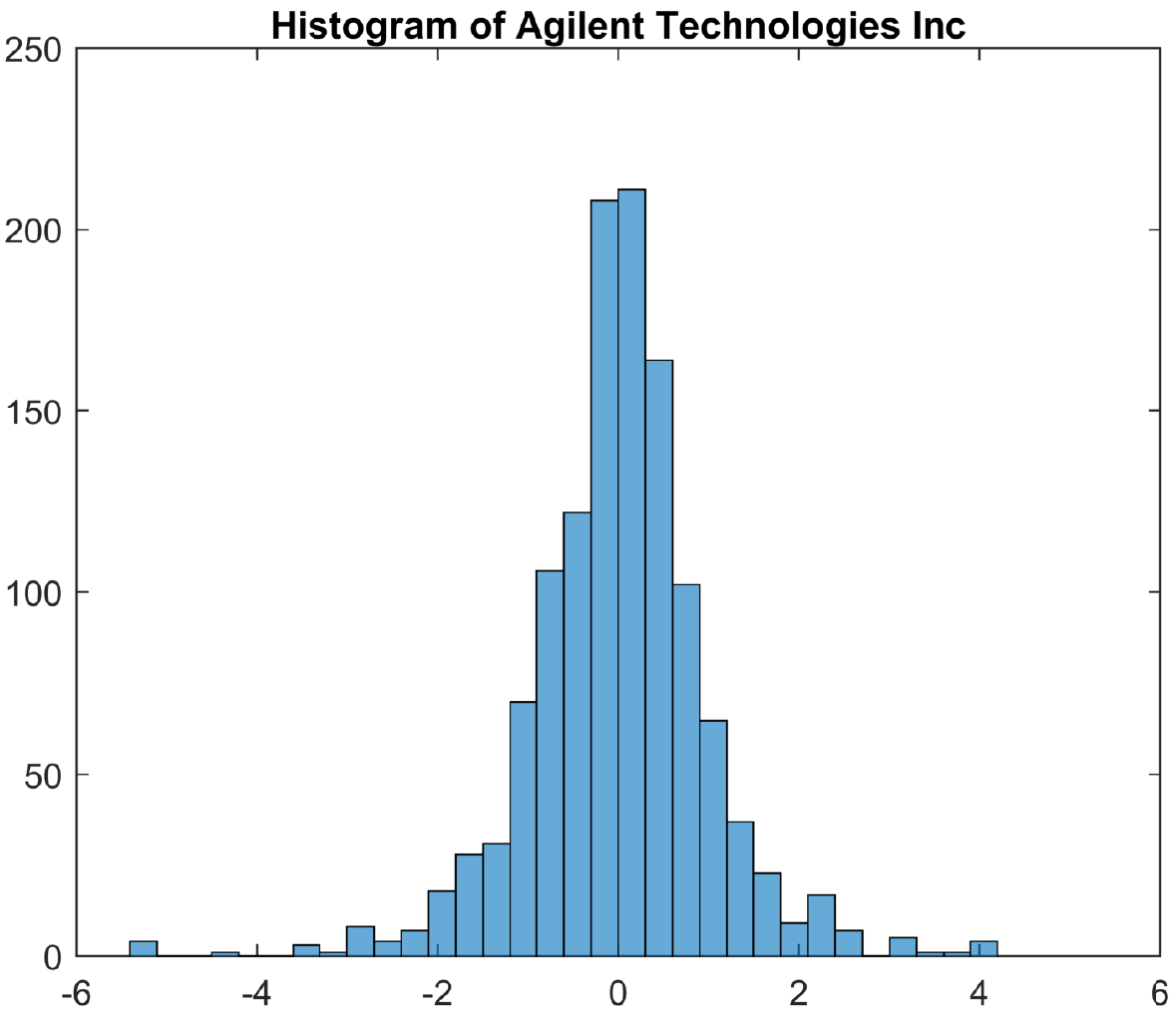}}
\caption{\footnotesize{We winsorize the daily return of $\log$ price so that all samples are within five   winsorized standard deviation from the winsorized mean. After preprocessing, we show histograms of the daily returns.}}
\label{fig:stock_hist_info}
\end{figure}


\begin{figure}[t]
\centering
\subfloat[Adobe Systems Inc from sector Information Technology.]{
\includegraphics[width=.45\columnwidth]{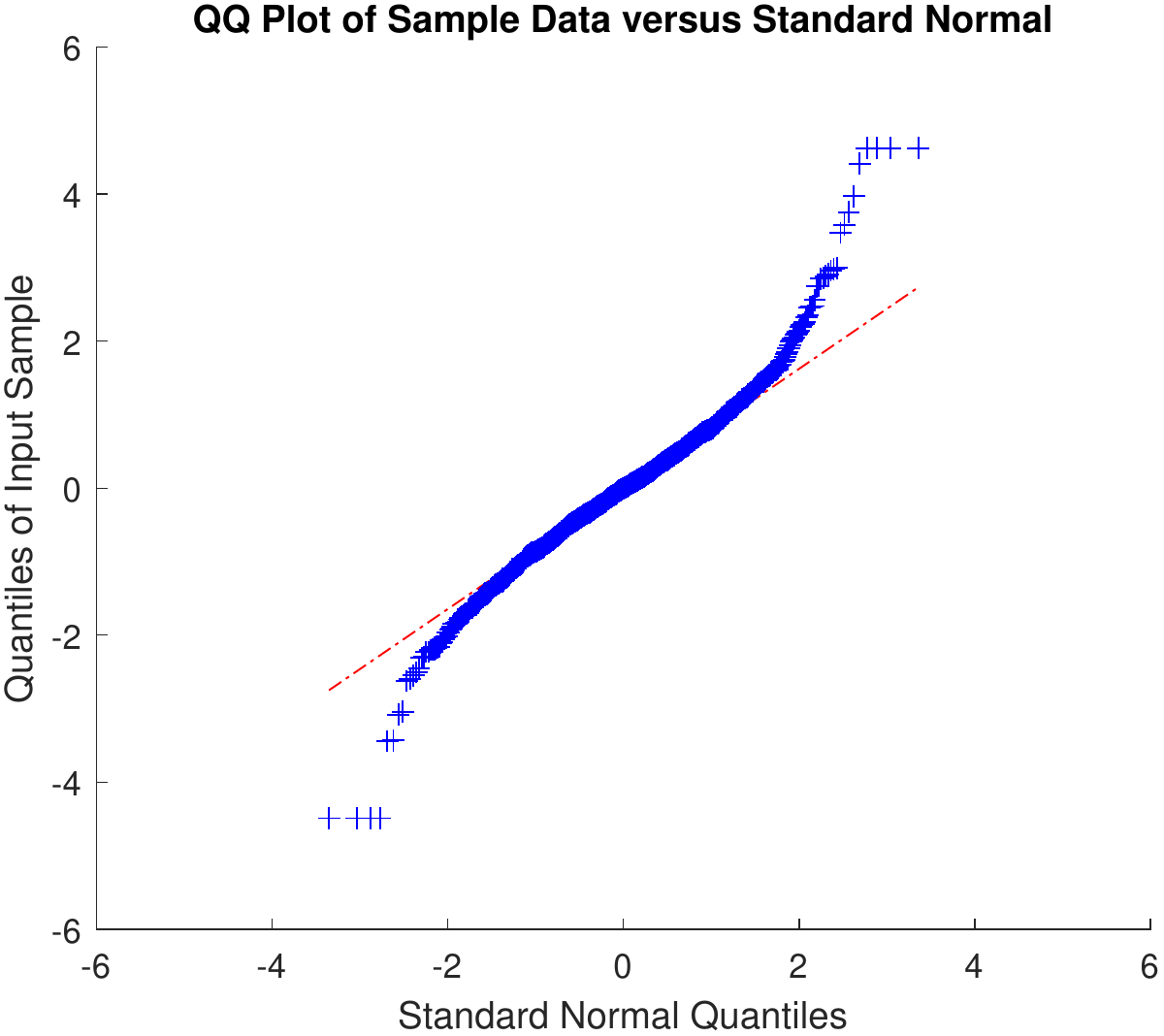}} $\quad\quad$
\subfloat[Agilent Technologies Inc from sector Information Technology.]{
\includegraphics[width=.45\columnwidth]{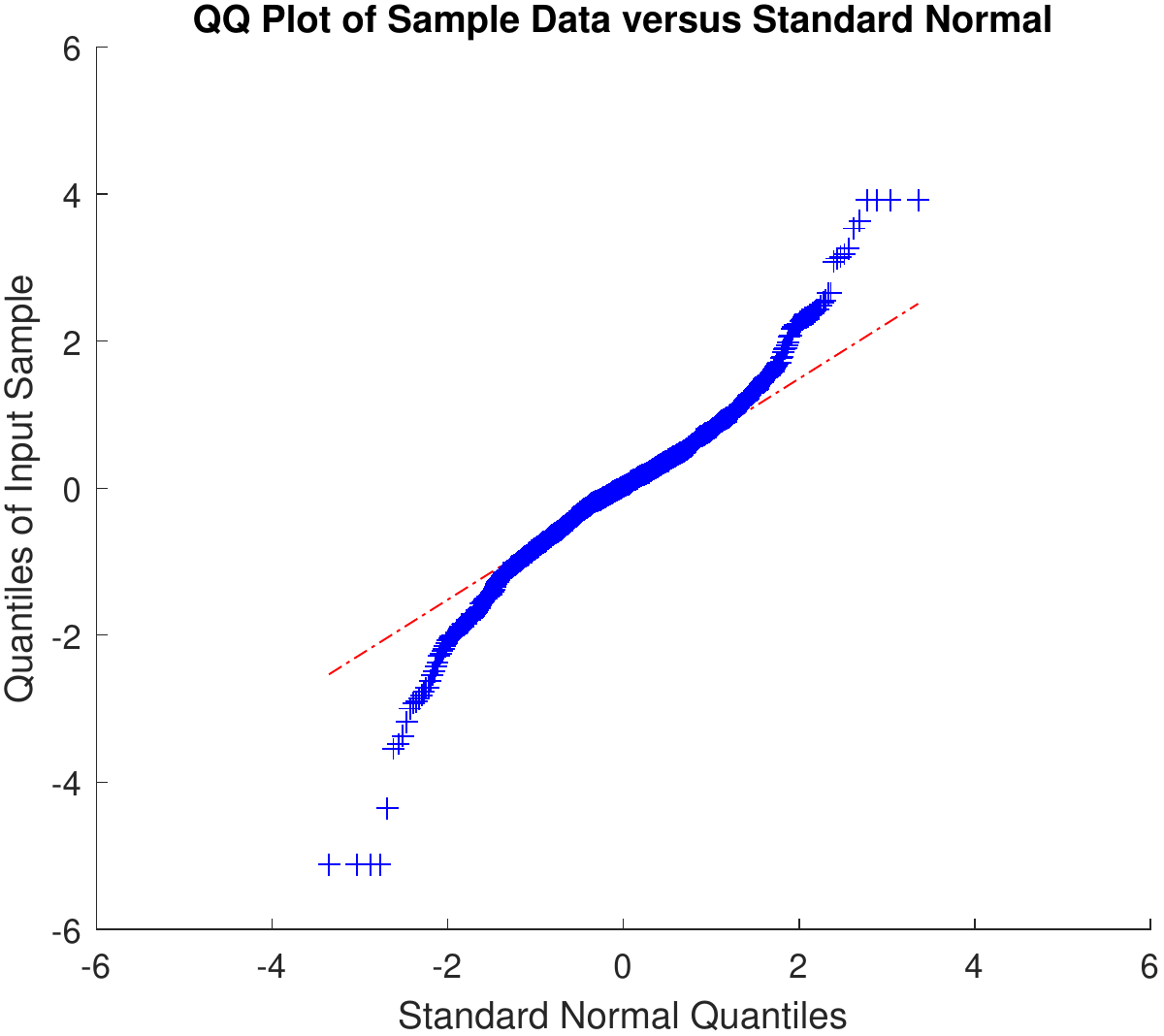}}
\caption{\footnotesize{After the same preprocessing, we present the QQ plot of the daily returns versus standard normal.}}
\label{fig:stock_qq_info}
\end{figure}


Here, we present details of the experiment using US equities data \cite{HugePackage}.  
We preprocess it by taking log-transformation and calculate the 
corresponding daily returns.  Obvious outliers are removed by winsorizing each variable so
that all samples are within five times the winsorized standard deviation from the winsorized mean.
After  preprocessing, we present example histograms and QQ plots from the Information Technology sector. 
In \Cref{fig:stock_hist_info} , we list the histograms of two typical companies in
this sector. As we can see from \Cref{fig:stock_qq_info}, even after preprocessing
on these stock prices, they are still highly non-normal and heavy tailed. 
We do not add any manual outliers as financial data is already heavy tailed and have many outliers \cite{de2018advances}. 
We also compare \Cref{alg:graphical_IHT} with the baseline NS approach (without  consideration for corruptions or outliers). 

\begin{figure}[t]
\centering
\subfloat[][Graph estimated by Robust NS
 (\Cref{alg:graphical_IHT}).]{
\includegraphics[width=.45\columnwidth]{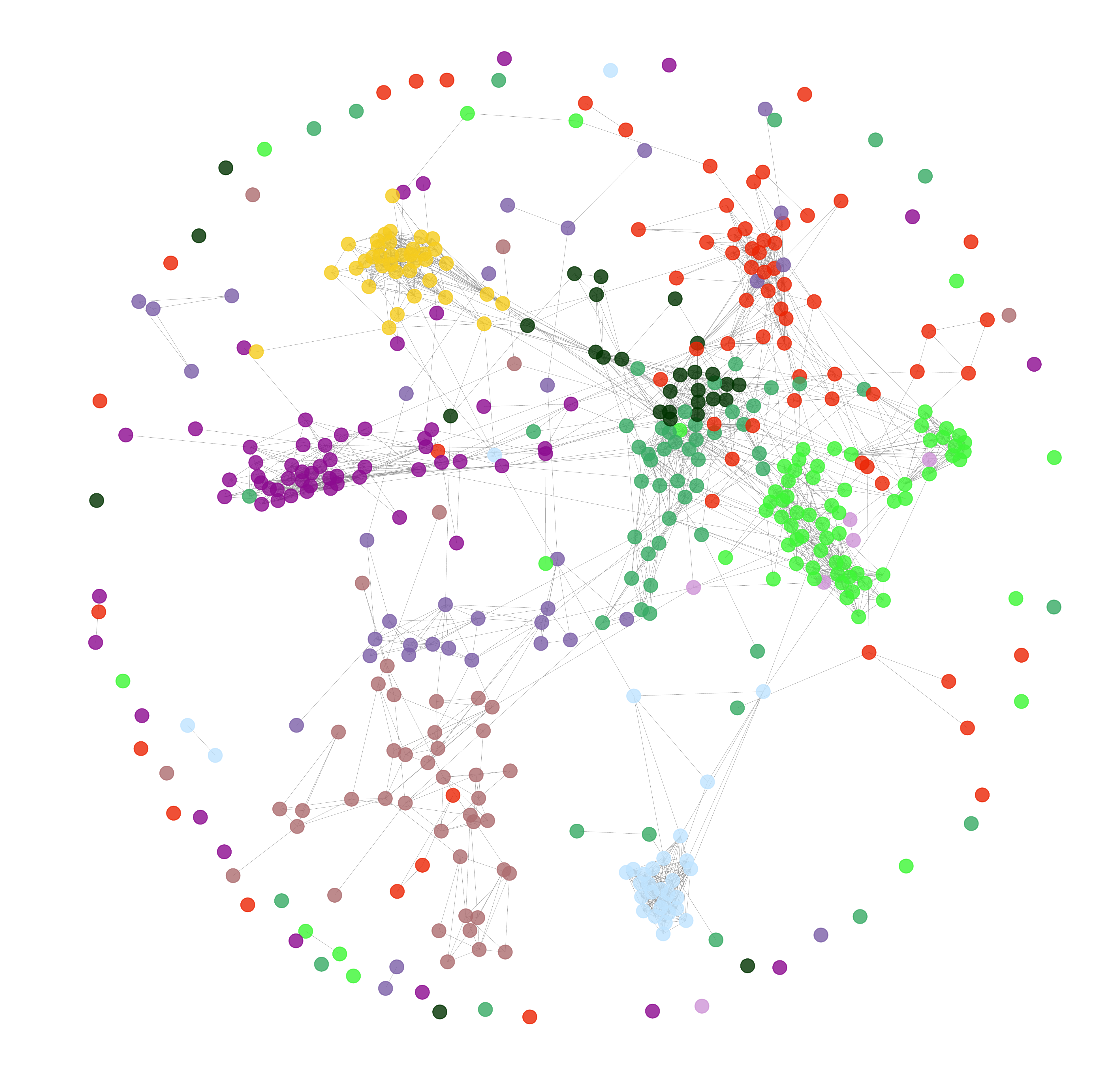}}
\subfloat[][Graph estimated
 by Vanilla NS approach.]{
\includegraphics[width=.45\columnwidth]{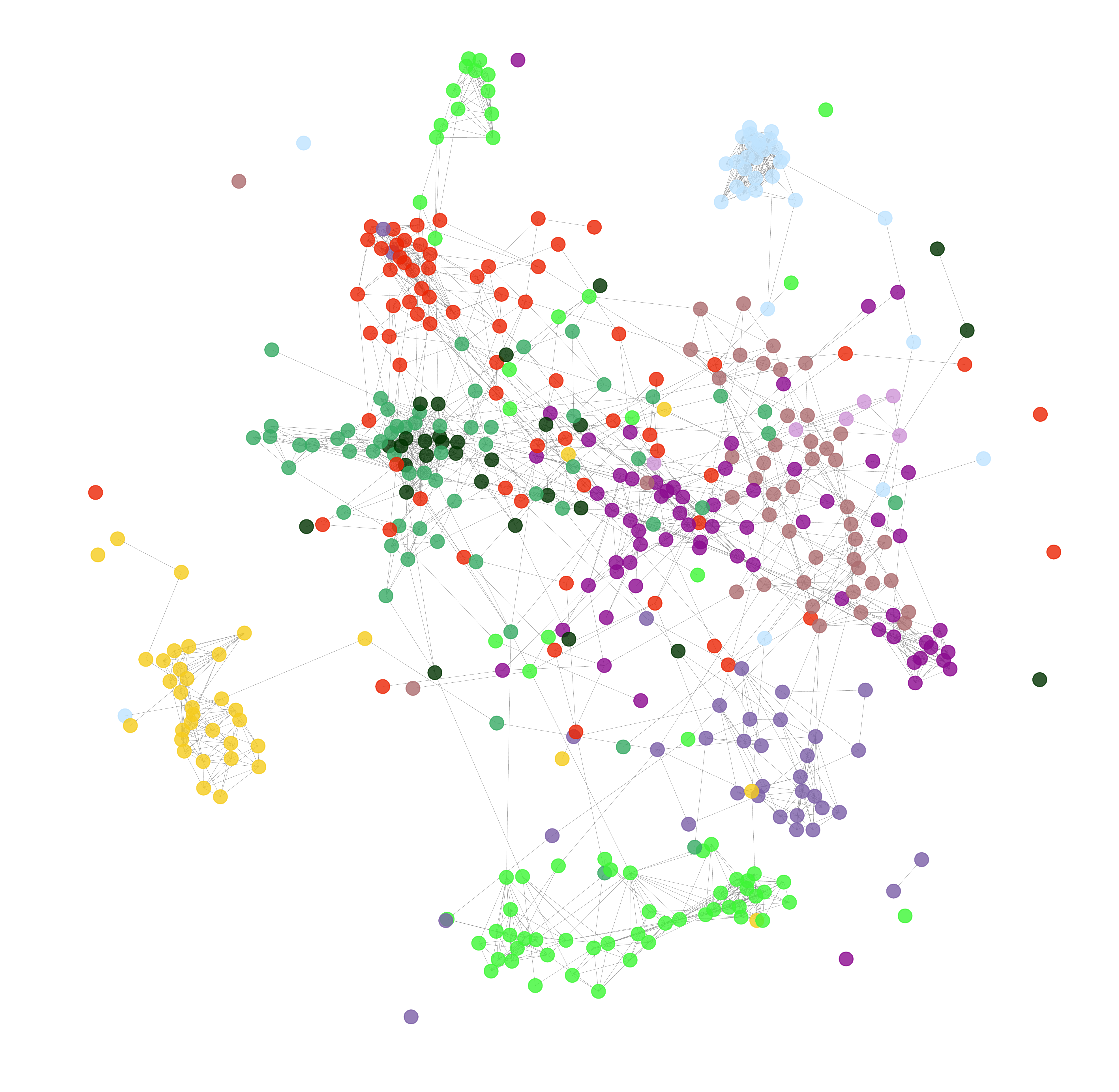}}
\caption{\footnotesize{
Graph estimated from the S\&P 500 stock data by \Cref{alg:graphical_IHT} and Vanilla NS approach. Variables are colored according to their sector. In particular, the stocks from sector Information Technology are colored as purple.}}
\label{fig:additional_clustering_appendix}
\end{figure}

We limit the number of edges to 2,000 for both methods. 
The cluster colored by purple denotes the Information Technology sector. In \Cref{fig:additional_clustering_appendix}, we can easily separate different clusters by using Robust NS.
However, the Vanilla NS approach \emph{cannot} distinguish the sector Information Technology (purple).
Furthermore, we can observe that stocks from Information Technology (colored by purple) are much better clustered by  \Cref{alg:graphical_IHT}.

